\documentclass{article}

\usepackage{microtype}
\usepackage{graphicx}
\usepackage{subcaption}
\usepackage{booktabs}
\usepackage{multirow}
\usepackage{amsmath}
\usepackage{amssymb}
\usepackage{mathtools}
\usepackage{amsthm}

\usepackage[hidelinks]{hyperref}
\usepackage{xurl}
\usepackage[capitalize,noabbrev]{cleveref}

\usepackage{cpg_arxiv}

\title{Localizing Text Anonymization for Trustworthy AI:\\
Extending RAT-Bench to Malaysian Microdata and PII}
\date{}

\authors{
  David Hong Liang Chew\affmark{1},
  Zexi Yao\affmark{2},
  Nata\v{s}a Kr\v{c}o\affmark{2},
  Matthieu Meeus\affmark{2},
  Waqas Khalid Obeidy\affmark{1},
  Yves-Alexandre de Montjoye\affmark{2}
}

\affiliations{
  \affmark{1}\textit{Payments Network Malaysia Sdn Bhd (PayNet)} \\
  \affmark{2}\textit{Imperial College London} \\
  Correspondence: David Hong Liang Chew, \href{mailto:davidchewhl@gmail.com}{davidchewhl@gmail.com}
}

\begin{document}
\maketitle

\begin{abstract}
As governments and organizations adopt foundation models for public services and workplace productivity, sensitive citizen, employee, and administrative text may enter LLM workflows through inference, retrieval, fine-tuning, or local model development. We examine whether text anonymizers transfer to new deployment settings by extending RAT-Bench to Malaysia using local microdata, Malaysian PII formats, and culturally grounded transcripts in Malaysian English and Bahasa Malaysia. We evaluate NER- and LLM-based anonymizers using an LLM attacker that infers attributes from anonymized text, measuring both re-identification success and text utility. Across Malaysian English and Bahasa Malaysia, the two LLM anonymizers provide the strongest explicit privacy--utility trade-offs, reducing Easy/Hard re-identification risk to 23--29\% while preserving BLEU scores of 0.77--0.94. Non-LLM tools show sharper failure modes, either preserving utility while leaving high residual risk or reducing risk through severe over-redaction, especially in Bahasa Malaysia. These findings suggest that anonymizers should be evaluated against deployment-relevant languages, identifier formats, and cultural contexts before being relied on in public-sector, workplace, or local AI pipelines.
\end{abstract}

\section{Introduction}

Governments and organizations are rapidly adopting foundation models for public services and workplace productivity. In Malaysia, this is moving from policy to practice: the AI at Work 2.0 initiative provides Google Workspace with Gemini tools and training to up to 445{,}000 Malaysian public officers \citep{digital_ai_at_work_2025}. Malaysia has also issued public-sector AI adoption guidelines covering legal compliance, ethics principles, roles and responsibilities, adoption procedures, and AI risk management \citep{jdn_public_sector_ai_guidelines}. More broadly, Malaysia's national AI governance guidance emphasizes safe, ethical, and responsible AI development and use \citep{malaysia_aige}.

This creates a practical privacy problem across the AI lifecycle. In the near term, sensitive citizen, employee, or administrative text may be submitted to LLM systems during inference-time use, such as drafting, summarization, search, or document analysis. In the longer term, similar text may also enter retrieval corpora, fine-tuning data, or local model-development pipelines, where memorization and later extraction become additional concerns. Privacy guidance for LLMs warns that users may overshare personal or confidential information with chatbots, and that logs may retain user interactions \citep{edpb_llm_privacy_2025}. Prior work further shows that LLMs can infer personal attributes from ordinary text, and that models can memorize and reveal training examples \citep{carlini2021extracting,carlini2023quantifying,staab2024beyond}. PII detection and anonymization are therefore natural guardrails, but their effectiveness must be evaluated against re-identification rather than span removal alone.

RAT-Bench evaluates anonymization through this residual re-identification risk rather than span-level recall \citep{ratbench}. However, its demographic grounding, identifiers, and cultural context are primarily U.S.-centric and do not account for other geographies. Malaysia is a useful stress test for contextual generalization: Malaysian English differs from U.S.-centric benchmark text in names, addresses, geography, local phrasing, and cultural references, while many interactions also occur in BM. We therefore evaluate both Malaysian English and Bahasa Malaysia transcripts, leaving richer code-switching to future work.

\begin{figure}[t]
    \centering
    \includegraphics[width=0.625\columnwidth]{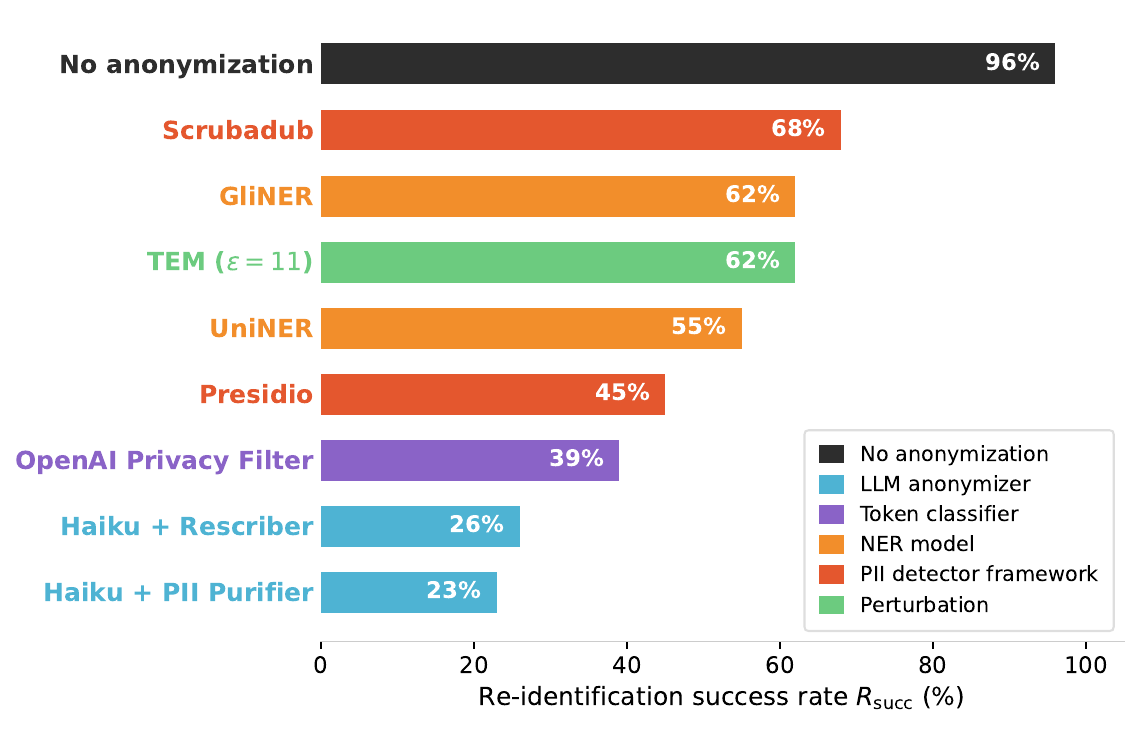}
    \caption{Malaysian English re-identification risk averaged over Easy and Hard settings.}
    \label{fig:intro-rsucc}
\end{figure}

We examine whether anonymizers retain their performance when transferred to the Malaysian setting. We make three contributions: (i) a framework for extending RAT-Bench to the Malaysian context using differentially private synthetic microdata derived from Malaysia IPUMS records; (ii) a localized PII-generation pipeline covering Malaysian names, addresses, postcodes, phone formats, National Registration Identity Card (NRIC)/passport numbers, Employee Provident Fund (EPF) numbers, Tax Identification Numbers (TINs), credit cards, and bank accounts; (iii) an evaluation of out-of-the-box anonymizers on Malaysian English and Bahasa Malaysia transcripts. We find that LLM-based anonymizers generally provide the strongest privacy--utility trade-off under unchanged prompts, while several non-LLM tools miss local PII, degrade on non-standard forms, or severely over-redact Bahasa Malaysia text. Together, these provide a case study of how privacy safeguards for AI deployment can fail when evaluated only on U.S.-centric identifiers and English-language assumptions.

\section{Background and Related Work}

\subsection*{LLM adoption and inference-time privacy.}

Public-sector and workplace LLM deployments create privacy risks even when models are not being trained or fine-tuned: prompts, uploaded documents, retrieved context, and conversation logs may contain sensitive personal or administrative information. Regulatory guidance highlights risks from users oversharing personal or confidential information with chatbots and from logs retaining user interactions \citep{edpb_llm_privacy_2025}. Research on privacy-preserving LLM inference also treats prompts themselves as sensitive inputs that may require protection when processed by cloud-hosted LLM services \citep{confidential_prompting}. These risks motivate practical guardrails such as PII detection and anonymization for inference, retrieval, and model-development workflows.

\subsection*{Memorization, attribute inference, and anonymization.}

Foundation models create privacy risks both by memorizing rare training strings and by inferring sensitive attributes from contextual cues. Training-data extraction and auditing work shows that language models can reproduce or reveal training examples under suitable attacks, motivating privacy controls before sensitive text is used in model-development pipelines \citep{carlini2019secret,carlini2021extracting,carlini2023quantifying,shi2024detecting}. Privacy leakage is not limited to memorization: LLMs can infer personal attributes from ordinary user-written text \citep{staab2024beyond}, and LLM agents can deanonymize pseudonymous users from unstructured online text at scale \citep{lermen2026largescale}. Text anonymization is therefore a practical mitigation, but its effectiveness should be evaluated against adversarial re-identification rather than span removal alone.

\subsection*{Risk-based evaluation and localization.}

Most de-identification and anonymization benchmarks evaluate protected-span detection or recall \citep{lison2021anonymisation,pilan2022tab,kim2024generalizing}. Recent work evaluates anonymization under stronger LLM-based adversaries and shows that LLM-based anonymization can improve privacy--utility trade-offs \citep{staab2025advanced,yang2025robust}. However, span removal alone does not measure whether the remaining text still contains enough direct or indirect information to identify an individual. This connects text anonymization to classical disclosure-control work, where risk is population-relative and depends on the uniqueness of indirect identifiers \citep{sweeney2002kanonymity,narayanan2008robust,demontjoye2013unique,rocher2019estimating,gadotti2024anonymization}. As such, risk is population-relative, conclusions from a U.S.-grounded benchmark such as RAT-Bench need not transfer unchanged to Malaysia \citep{ratbench}. We therefore test whether RAT-Bench's findings remain stable after localizing the reference microdata, direct identifiers, and language context.

\section{Malaysian Benchmark Construction}

\subsection{Source Microdata and Record Sampling}
We use year-2000 Malaysian IPUMS microdata, the latest Malaysian microdata available to us. The original data contains more than 400,000 records and cannot be publicly released. We select and transform relevant IPUMS variables to align with the identifier categories defined in RAT-Bench, then use PrivBayes via Reprosyn with privacy budget $\epsilon=5$ to generate a synthetic microdata sample of 400,000 records \citep{zhang2017privbayes,reprosyn}. This synthetic microdata is used for benchmark construction and re-identification scoring. 

Following RAT-Bench, we select benchmark records that are already high-risk before anonymization. For a candidate record, baseline correctness measures how likely an attacker is to recover the correct individual from the reference population using a sampled set of indirect identifiers. We sample 100 high-risk benchmark records by selecting records whose baseline correctness under five sampled indirect identifiers exceeds $\theta_0=0.9$.

\subsection{Indirect Identifier Construction and Enrichment}
We map Malaysian IPUMS variables to RAT-Bench's nine indirect-identifier categories: state, sex, age, race, marital status, education, employment status, occupation, and citizenship. Two mappings require approximation: place of birth is used as the closest available proxy for state of residence, and education level is combined with highest certificate obtained to approximate RAT-Bench's finer-grained education attribute. Full mapping details are in Appendix~\ref{app:ipums-mapping}.

Postcode is needed for two parts of the benchmark: generating realistic Malaysian addresses and scoring partial address recovery during re-identification evaluation. However, the Malaysia IPUMS extract does not provide postcode-level geography. We therefore enrich the synthetic microdata with state-consistent postcodes using a population-weighted procedure, rather than sampling postcodes uniformly within each state. Unlike RAT-Bench's U.S. setting, where ZIP codes can be mapped to PUMA geography, our Malaysian setting has no PUMA-equivalent field, so we use postcodes directly in re-identification scoring.

For each record, we first sample a district \(d\) within the record's state \(s\), proportional to district population \citep{opendosm_population_district}:

\[
P(d \mid s)
=
\frac{\mathrm{population}(d)}
{\sum_{\tilde{d} \in \mathcal{D}(s)} \mathrm{population}(\tilde{d})},
\]
where \(\mathcal{D}(s)\) is the set of districts in state \(s\). We then sample a postcode uniformly from the postcode pool matched to that district \citep{malaysia_postcodes}:
\[
P(z \mid d)
=
\frac{1}{|\mathcal{Z}_d|},
\]
where \(z\) is the sampled postcode and \(\mathcal{Z}_d\) is the set of postcodes matched to district \(d\). Additional postcode-matching details are provided in Appendix~\ref{app:postcode}.

\subsection{Malaysian Direct Identifier Generation}

We also introduce three Malaysian-relevant identifiers: bank account number, EPF number, and TIN. These are strongly identifying locally: bank account numbers and mobile numbers can reveal or confirm account-holder names in common payment workflows, while EPF and TIN are unique administrative identifiers linked to an individual's NRIC.

Identifiers are generated using local format constraints. Names are conditioned on state, ethnicity, and gender; addresses are synthetic but follow Malaysian address structure and valid postcodes; phone numbers include mobile and landline formats; and NRICs encode date of birth, state code, and gender. To ensure coverage and comparability, we stratify direct identifier assignment: in each 100-record explicit setting, each of the nine direct identifier categories appears 11 times, with one category appearing 12 times. Full generation rules and format constraints are provided in Appendix~\ref{app:direct-identifiers}.

\subsection{Transcript Generation and Quality Control}
Following RAT-Bench, we generate transcripts in three scenarios: medical consultation, chatbot conversation, and meeting transcript. We repeat generation in Malaysian English and Bahasa Malaysia, grounding both in Malaysian names, geography, institutions, and social context. For each language and difficulty setting, we generate 100 transcripts from the same selected high-risk records.

We use RAT-Bench's three difficulty settings. Easy states one direct identifier and five indirect identifiers explicitly; Hard keeps them explicit but uses non-standard forms; and Implicit contains only indirect identifiers, conveyed through contextual clues. Implicit is less straightforward in the Malaysian context because target attributes must be recoverable from indirect local cues. Initial generations did not consistently contain enough evidence for the attacker to infer the target attributes. We therefore manually corrected 61/100 Malaysian English Implicit transcripts to strengthen clue bundles while keeping target values implicit, increasing no-anonymization re-identification from 5\% to 35\%, close to RAT-Bench's 32\% Implicit risk. Bahasa Malaysia Implicit showed the same issue, so we treat it as exploratory and exclude it from the main comparison.

\section{Experimental Setup}

\begin{center}
    \includegraphics[width=\columnwidth]{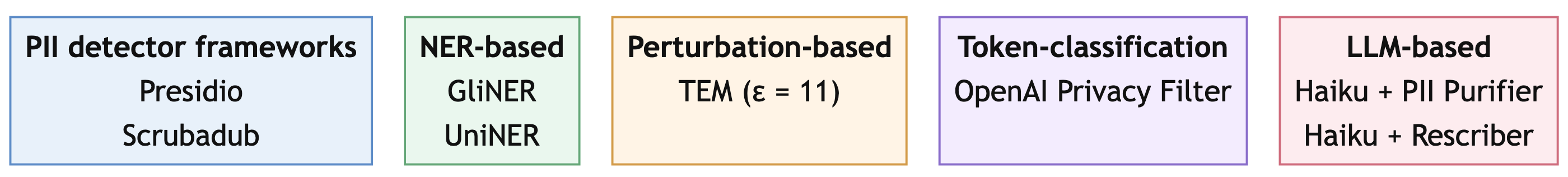}
    \captionof{figure}{Anonymizer families evaluated in this work.}
    \label{fig:anonymizer-taxonomy}
\end{center}

\subsection*{Anonymizers.}

We evaluate seven anonymizers from RAT-Bench: five non-LLM methods (Presidio, Scrubadub, GliNER, UniNER, TEM with \(\epsilon=11\)) and two LLM-based methods using Claude Haiku 4.5 with the PII Purifier and Rescriber prompts. We add OpenAI Privacy Filter \citep{openai_privacy_filter}. All anonymizers are used out of the box.

\subsection*{Generator and attacker.}

We use Claude Sonnet 4.5 for transcript generation and attack. This provides some model variation across the pipeline while retaining a strong generator and attacker. The attacker prompt follows RAT-Bench, with one localized addition instructing the attacker to make inferences from a Malaysian context and to use 1 January 2000 as the reference date for time-based reasoning. The exact modified prompt footer is provided in Appendix~\ref{app:attacker-prompt}.

\subsection*{Metrics.}

Following RAT-Bench, a record is considered re-identified if at least one direct identifier is recovered, or if the estimated re-identification risk from recovered indirect identifiers exceeds \(\theta=0.2\). This threshold corresponds to the \(k=5\) intuition from \(k\)-anonymity, where a record should remain indistinguishable from at least four others under quasi-identifier matching \citep{sweeney2002kanonymity}. We report re-identification success rate, with lower values indicating lower residual privacy risk, and BLEU between the original and anonymized text, with higher values indicating better utility.

\section{Results}

\begin{table*}[t]

\caption{Main results for Malaysian English and Bahasa Malaysia. $R_{succ}$ is re-identification success rate. Avg. is computed over Easy and Hard settings. Bahasa Malaysia Implicit columns are omitted because the no-anonymization result remains too low at 7\% to support meaningful comparison.}

\label{tab:main-results}
\centering
\begin{small}
\setlength{\tabcolsep}{5.5pt}
\begin{tabular}{lrrrrrrrrrr}
\toprule
\multirow{3}{*}{Method}
& \multicolumn{6}{c}{Malaysian English}
& \multicolumn{4}{c}{Bahasa Malaysia} \\
\cmidrule(lr){2-7}\cmidrule(lr){8-11}
& \multicolumn{4}{c}{$R_{succ}$ (\% $\downarrow$)}
& \multicolumn{2}{c}{BLEU ($\uparrow$)}
& \multicolumn{3}{c}{$R_{succ}$ (\% $\downarrow$)}
& \multicolumn{1}{c}{BLEU ($\uparrow$)} \\
\cmidrule(lr){2-5}\cmidrule(lr){6-7}
\cmidrule(lr){8-10}\cmidrule(l){11-11}
& Easy & Hard & Avg. & Implicit & Avg. & Implicit
& Easy & Hard & Avg. & Avg. \\
\midrule
No Anonymization & 100 & 91 & 96 & 35 & 1.00 & 1.00 & 100 & 96 & 98 & 1.00 \\
Presidio & 40 & 50 & 45 & 2 & 0.82 & 0.85 & 52 & \textbf{28} & 40 & 0.17 \\
Scrubadub & 64 & 72 & 68 & 11 & 0.77 & 0.82 & 68 & 61 & 65 & 0.41 \\
GliNER & 66 & 57 & 62 & 2 & 0.87 & 0.66 & 66 & 63 & 65 & 0.49 \\
UniNER & 54 & 56 & 55 & 25 & 0.65 & 0.96 & 51 & 73 & 62 & 0.87 \\
TEM $\epsilon=11$ & 55 & 69 & 62 & 30 & 0.65 & 0.70 & 45 & 76 & 61 & 0.25 \\
Haiku 4.5 + PII Purifier Prompt & 21 & \textbf{24} & \textbf{23} & 7 & \textbf{0.94} & 0.97 & 20 & 37 & 29 & 0.92 \\
Haiku 4.5 + Rescriber Prompt & 21 & 31 & 26 & \textbf{1} & 0.83 & 0.87 & \textbf{8} & 37 & \textbf{23} & 0.77 \\
OpenAI Privacy Filter & \textbf{18} & 59 & 39 & 18 & \textbf{0.94} & \textbf{0.98} & 42 & 83 & 63 & \textbf{0.94} \\
\bottomrule
\end{tabular}
\end{small}
\end{table*}

\begin{figure*}[t]
    \centering
    \begin{subfigure}[t]{0.48\textwidth}
        \centering
        \includegraphics[width=\linewidth]{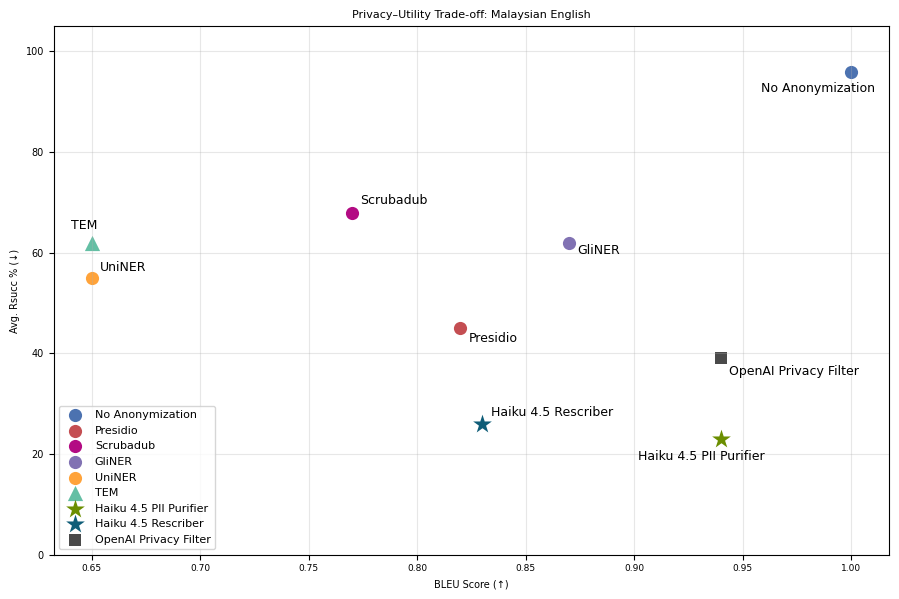}
        \caption{Malaysian English}
        \label{fig:tradeoff-english}
    \end{subfigure}
    \hfill
    \begin{subfigure}[t]{0.48\textwidth}
        \centering
        \includegraphics[width=\linewidth]{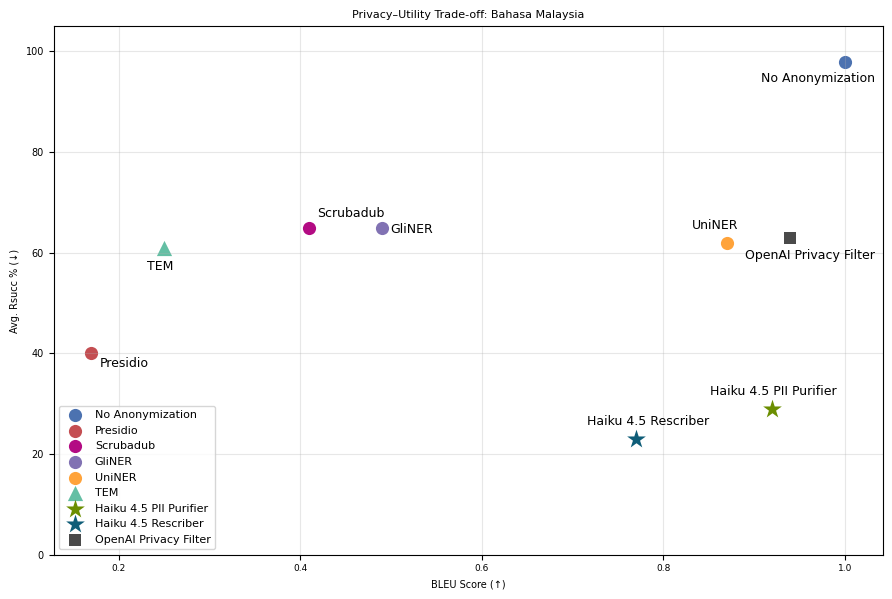}
        \caption{Bahasa Malaysia}
        \label{fig:tradeoff-bm}
    \end{subfigure}
    \caption{Privacy--utility trade-off across anonymizers, using Easy/Hard average re-identification success rate and BLEU values from \cref{tab:main-results}. Lower \(R_{succ}\) indicates stronger privacy protection, while higher BLEU indicates higher text utility.}
    \label{fig:privacy-utility}
\end{figure*}

\cref{tab:main-results} shows three main findings. First, LLM-based anonymizers provide the strongest Easy/Hard privacy--utility trade-off in Malaysian English, as visualized in \cref{fig:privacy-utility}. Haiku + PII Purifier achieves the lowest average explicit re-identification rate (23\%) while preserving high utility (BLEU 0.94). Haiku + Rescriber is close in privacy (26\%) but has lower utility (BLEU 0.83). OpenAI Privacy Filter preserves utility similarly well, but its risk increases sharply from Easy to Hard identifiers (18\% to 59\%), suggesting weaker protection for non-standard local forms. Among non-LLM methods, Presidio is strongest but still leaves 45\% average explicit risk.

Second, the Malaysian English Implicit setting evaluates contextual re-identification: whether indirect cues can still identify someone when direct identifiers are absent. Since no direct identifiers are present, record-level \(R_{succ}\) depends entirely on recovered indirect attributes. Haiku + Rescriber performs best in this setting, reducing \(R_{succ}\) to 1\% while preserving BLEU 0.87. Presidio is the strongest non-LLM method on Implicit risk (2\%), but with lower utility (BLEU 0.85) than Haiku + Rescriber and Haiku + PII Purifier. Haiku + PII Purifier has higher residual risk (7\%) but preserves the most utility among these methods (BLEU 0.97).

Third, Bahasa Malaysia shows a larger gap between LLM-based and non-LLM privacy--utility trade-offs. The LLM methods remain comparatively stable: Haiku + Rescriber achieves the lowest average explicit risk (23\%), while Haiku + PII Purifier preserves higher utility with moderately higher risk (29\%, BLEU 0.92). In contrast, several non-LLM methods show sharper failure modes. Presidio reduces average explicit risk to 40\%, but BLEU drops to 0.17, indicating severe over-redaction. Conversely, OpenAI Privacy Filter preserves utility well (BLEU 0.94) but leaves high residual risk (63\%). We provide an example of such over-redaction in Appendix~\ref{app:bm-failure-example}.

Overall, non-LLM tools exhibit mixed failure modes: some preserve utility but leave high residual risk, while others heavily redact benign text yet can still leave local PII unredacted. Relative to RAT-Bench \citep{ratbench}, our results preserve the broad LLM advantage but show a clearer gap between LLM- and NER-based method performance in Malaysian English and Bahasa Malaysia across both Easy and Hard examples.

\section{Discussion and Limitations}
\label{sec:limitations}

Our results support RAT-Bench's conclusion that LLM-based anonymizers often provide a stronger privacy--utility trade-off than non-LLM systems, but show that U.S.-centric benchmarks can understate deployment failures elsewhere. In Malaysia, local identifier formats, language, and cultural context make failures more visible: some non-LLM tools miss locally meaningful identifiers or non-standard explicit forms, while in Bahasa Malaysia they can over-redact large portions of text and still miss direct identifiers. This suggests that anonymizers should be trained and evaluated on deployment-relevant languages, identifier formats, and cultural contexts, not only on U.S.-centric data.

For public-sector and workplace AI deployments, anonymizer choice should be guided by the language and context of deployment rather than English or U.S.-centric benchmarks alone. LLM-based anonymizers appear more adaptable under unchanged anonymization prompts, but are costlier and should still be evaluated in the target context. This is especially important when public officers or employees use LLM tools for drafting, summarization, search, or document analysis, where prompts and retrieved context may contain citizen, employee, or administrative information. The same concern extends to retrieval corpora, fine-tuning data, and local model-development pipelines, where retained PII can create downstream memorization or extraction risks. A tool that performs well on standard English PII may still miss local identifiers such as NRIC-related formats, EPF, TIN, bank-account numbers, state-dependent landlines, or Malaysian address forms. Conversely, overly aggressive redaction may reduce the usefulness of text for service improvement, internal knowledge systems, or local model development. Localized privacy evaluation is therefore part of responsible AI adoption, not merely a benchmark exercise.

\subsection*{Limitations.}

Due to data-use restrictions, we cannot release the original or synthetic Malaysia IPUMS microdata or generated benchmark. State is approximated using place of birth rather than current residence. Postcode enrichment uses year 2020 district population statistics, the earliest available source we found, although the base microdata is from 2000. Implicit generation remains challenging: Malaysian English Implicit examples required quality control, while Bahasa Malaysia Implicit remains preliminary. Malaysia-specific recognizers or fine-tuned systems could improve non-LLM tools; our goal is to test default cross-context generalization.

\section{Conclusion}

We extend RAT-Bench to Malaysia by constructing differentially private synthetic microdata derived from Malaysian IPUMS records, adding Malaysian-specific direct identifiers, and evaluating anonymizers on Malaysian English and Bahasa Malaysia transcripts. The broad RAT-Bench pattern transfers: LLM-based anonymizers generally provide the strongest privacy--utility trade-off. However, the Malaysian setting makes several failure modes more visible, including missed local identifiers, weaker protection for non-standard explicit forms, and severe over-redaction in Bahasa Malaysia text. As governments and organizations adopt LLMs in public-sector and workplace settings, anonymization benchmarks should be localized across identifier formats, language use, and cultural context before their results are used to justify privacy controls for sensitive local text.

\section*{Impact Statement}
This paper studies text anonymization as a privacy safeguard for localized and multilingual foundation-model use. Its potential positive impact is to support safer public-sector, workplace, and local AI deployments by showing that anonymization tools should be evaluated against local identifiers, languages, and cultural context before they are relied on as privacy controls. This matters both for inference-time use, where prompts and documents may contain sensitive information, and for model-development workflows, where retained PII can create downstream memorization or extraction risks. A potential negative impact is that reporting anonymizer failures could help adversaries probe deployed systems; we mitigate this by reporting aggregate findings and not releasing restricted microdata or generated benchmark text. Overall, we argue that localized privacy evaluation is part of responsible AI adoption across societies and languages.

\bibliography{references}

@misc{ratbench,
      title={RAT-Bench: A Comprehensive Benchmark for Text Anonymization}, 
      author={Nata{\v{s}}a Kr{\v{c}}o and Zexi Yao and Matthieu Meeus and Yves-Alexandre de Montjoye},
      year={2026},
      eprint={2602.12806},
      archivePrefix={arXiv},
      primaryClass={cs.CL},
      url={https://arxiv.org/abs/2602.12806}, 
}

@inproceedings{carlini2019secret,
  title = {The Secret Sharer: Evaluating and Testing Unintended Memorization in Neural Networks},
  author = {Carlini, Nicholas and Liu, Chang and Erlingsson, {\'U}lfar and Kos, Jernej and Song, Dawn},
  booktitle = {28th USENIX Security Symposium (USENIX Security 19)},
  pages = {267--284},
  year = {2019},
  address = {Santa Clara, CA},
  publisher = {USENIX Association},
  month = aug,
  url = {https://www.usenix.org/conference/usenixsecurity19/presentation/carlini}
}

@inproceedings{carlini2021extracting,
  title = {Extracting Training Data from Large Language Models},
  author = {Carlini, Nicholas and Tram{\`e}r, Florian and Wallace, Eric and Jagielski, Matthew and Herbert-Voss, Ariel and Lee, Katherine and Roberts, Adam and Brown, Tom and Song, Dawn and Erlingsson, {\'U}lfar and Oprea, Alina and Raffel, Colin},
  booktitle = {30th USENIX Security Symposium},
  pages = {2633--2650},
  year = {2021}
}

@inproceedings{
carlini2023quantifying,
title={Quantifying Memorization Across Neural Language Models},
author={Nicholas Carlini and Daphne Ippolito and Matthew Jagielski and Katherine Lee and Florian Tramer and Chiyuan Zhang},
booktitle={The Eleventh International Conference on Learning Representations },
year={2023},
url={https://openreview.net/forum?id=TatRHT_1cK}
}

@inproceedings{
shi2024detecting,
title={Detecting Pretraining Data from Large Language Models},
author={Weijia Shi and Anirudh Ajith and Mengzhou Xia and Yangsibo Huang and Daogao Liu and Terra Blevins and Danqi Chen and Luke Zettlemoyer},
booktitle={The Twelfth International Conference on Learning Representations},
year={2024},
url={https://openreview.net/forum?id=zWqr3MQuNs}
}

@inproceedings{
staab2024beyond,
title={Beyond Memorization: Violating Privacy via Inference with Large Language Models},
author={Robin Staab and Mark Vero and Mislav Balunovic and Martin Vechev},
booktitle={The Twelfth International Conference on Learning Representations},
year={2024},
url={https://openreview.net/forum?id=kmn0BhQk7p}
}

@inproceedings{
staab2025advanced,
title={Language Models are Advanced Anonymizers},
author={Robin Staab and Mark Vero and Mislav Balunovic and Martin Vechev},
booktitle={The Thirteenth International Conference on Learning Representations},
year={2025},
url={https://openreview.net/forum?id=82p8VHRsaK}
}

@inproceedings{yang2025robust,
    title = "Robust Utility-Preserving Text Anonymization Based on Large Language Models",
    author = "Yang, Tianyu  and
      Zhu, Xiaodan  and
      Gurevych, Iryna",
    editor = "Che, Wanxiang  and
      Nabende, Joyce  and
      Shutova, Ekaterina  and
      Pilehvar, Mohammad Taher",
    booktitle = "Proceedings of the 63rd Annual Meeting of the Association for Computational Linguistics (Volume 1: Long Papers)",
    month = jul,
    year = "2025",
    address = "Vienna, Austria",
    publisher = "Association for Computational Linguistics",
    url = "https://aclanthology.org/2025.acl-long.1404/",
    doi = "10.18653/v1/2025.acl-long.1404",
    pages = "28922--28941",
    ISBN = "979-8-89176-251-0",
    
}

@inproceedings{lison2021anonymisation,
    title = "Anonymisation Models for Text Data: State of the art, Challenges and Future Directions",
    author = "Lison, Pierre  and
      Pil{\'a}n, Ildik{\'o}  and
      Sanchez, David  and
      Batet, Montserrat  and
      {\O}vrelid, Lilja",
    editor = "Zong, Chengqing  and
      Xia, Fei  and
      Li, Wenjie  and
      Navigli, Roberto",
    booktitle = "Proceedings of the 59th Annual Meeting of the Association for Computational Linguistics and the 11th International Joint Conference on Natural Language Processing (Volume 1: Long Papers)",
    month = aug,
    year = "2021",
    address = "Online",
    publisher = "Association for Computational Linguistics",
    url = "https://aclanthology.org/2021.acl-long.323/",
    doi = "10.18653/v1/2021.acl-long.323",
    pages = "4188--4203",
    
}

@article{pilan2022tab,
    title = "The Text Anonymization Benchmark ({TAB}): A Dedicated Corpus and Evaluation Framework for Text Anonymization",
    author = "Pil{\'a}n, Ildik{\'o}  and
      Lison, Pierre  and
      {\O}vrelid, Lilja  and
      Papadopoulou, Anthi  and
      S{\'a}nchez, David  and
      Batet, Montserrat",
    journal = "Computational Linguistics",
    volume = "48",
    number = "4",
    month = dec,
    year = "2022",
    address = "Cambridge, MA",
    publisher = "MIT Press",
    url = "https://aclanthology.org/2022.cl-4.19/",
    doi = "10.1162/coli_a_00458",
    pages = "1053--1101",
    
}

@inproceedings{kim2024generalizing,
    title = "Generalizing Clinical De-identification Models by Privacy-safe Data Augmentation using {GPT}-4",
    author = "Kim, Woojin  and
      Hahm, Sungeun  and
      Lee, Jaejin",
    editor = "Al-Onaizan, Yaser  and
      Bansal, Mohit  and
      Chen, Yun-Nung",
    booktitle = "Proceedings of the 2024 Conference on Empirical Methods in Natural Language Processing",
    month = nov,
    year = "2024",
    address = "Miami, Florida, USA",
    publisher = "Association for Computational Linguistics",
    url = "https://aclanthology.org/2024.emnlp-main.1181/",
    doi = "10.18653/v1/2024.emnlp-main.1181",
    pages = "21204--21218",
    
}

@article{sweeney2002kanonymity,
author = {Sweeney, Latanya},
title = {k-anonymity: a model for protecting privacy},
year = {2002},
issue_date = {October 2002},
publisher = {World Scientific Publishing Co., Inc.},
address = {USA},
volume = {10},
number = {5},
issn = {0218-4885},
url = {https://doi.org/10.1142/S0218488502001648},
doi = {10.1142/S0218488502001648}
}

@article{rocher2019estimating,
  author  = {Rocher, Luc and Hendrickx, Julien M. and de Montjoye, Yves-Alexandre},
  title   = {Estimating the success of re-identifications in incomplete datasets using generative models},
  journal = {Nature Communications},
  volume  = {10},
  number  = {3069},
  year    = {2019},
  doi     = {10.1038/s41467-019-10933-3},
  url     = {https://doi.org/10.1038/s41467-019-10933-3},
  date    = {2019-07-23}
}

@inproceedings{narayanan2008robust,
  author={Narayanan, Arvind and Shmatikov, Vitaly},
  booktitle={2008 IEEE Symposium on Security and Privacy (sp 2008)}, 
  title={Robust De-anonymization of Large Sparse Datasets}, 
  year={2008},
  volume={},
  number={},
  pages={111-125},
  doi={10.1109/SP.2008.33}}

@article{demontjoye2013unique,
  author  = {de Montjoye, Yves-Alexandre and Hidalgo, C{\'e}sar A. and Verleysen, Michel and Blondel, Vincent D.},
  title   = {Unique in the Crowd: The privacy bounds of human mobility},
  journal = {Scientific Reports},
  volume  = {3},
  number  = {1376},
  year    = {2013},
  doi     = {10.1038/srep01376},
  url     = {https://doi.org/10.1038/srep01376},
  date    = {2013-03-25}
}

@article{gadotti2024anonymization,
author = {Andrea Gadotti  and Luc Rocher  and Florimond Houssiau  and Ana-Maria Cre{\c{t}}u  and Yves-Alexandre de Montjoye },
title = {Anonymization: The imperfect science of using data while preserving privacy},
journal = {Science Advances},
volume = {10},
number = {29},
pages = {eadn7053},
year = {2024},
doi = {10.1126/sciadv.adn7053},
URL = {https://www.science.org/doi/abs/10.1126/sciadv.adn7053},
eprint = {https://www.science.org/doi/pdf/10.1126/sciadv.adn7053},
}

@article{zhang2017privbayes,
author = {Zhang, Jun and Cormode, Graham and Procopiuc, Cecilia M. and Srivastava, Divesh and Xiao, Xiaokui},
title = {PrivBayes: Private Data Release via Bayesian Networks},
year = {2017},
issue_date = {December 2017},
publisher = {Association for Computing Machinery},
address = {New York, NY, USA},
volume = {42},
number = {4},
issn = {0362-5915},
url = {https://doi.org/10.1145/3134428},
doi = {10.1145/3134428},
journal = {ACM Trans. Database Syst.},
month = oct,
articleno = {25},
numpages = {41}
}

@misc{reprosyn,
  author       = {{Alan Turing Institute}},
  title        = {{Reprosyn}: Synthesising Tabular Data},
  year         = {2024},
  howpublished = {\url{https://github.com/alan-turing-institute/reprosyn}},
  note         = {Python library for generating synthetic data. Accessed 2026-03-30}
}

@misc{opendosm_population_district,
  author       = {{Department of Statistics Malaysia}},
  title        = {Population Table: Administrative Districts},
  year         = {2024},
  howpublished = {\url{https://open.dosm.gov.my/data-catalogue/population_district}},
  note         = {OpenDOSM data catalogue. Accessed 2026-04-29}
}

@misc{malaysia_postcodes,
  author       = {{heiswayi}},
  title        = {Malaysia Postcodes},
  year         = {2025},
  howpublished = {\url{https://github.com/heiswayi/malaysia-postcodes}},
  note         = {GitHub repository. Accessed 2026-04-29}
}

@misc{openai_privacy_filter,
  author       = {{OpenAI}},
  title        = {{OpenAI Privacy Filter Model Card}},
  year         = {2026},
  month        = apr,
  howpublished = {\url{https://cdn.openai.com/pdf/c66281ed-b638-456a-8ce1-97e9f5264a90/OpenAI-Privacy-Filter-Model-Card.pdf}},
  note         = {Accessed 2026-05-08}
}

@misc{lermen2026largescale,
      title={Large-scale online deanonymization with LLMs}, 
      author={Simon Lermen and Daniel Paleka and Joshua Swanson and Michael Aerni and Nicholas Carlini and Florian Tramèr},
      year={2026},
      eprint={2602.16800},
      archivePrefix={arXiv},
      primaryClass={cs.CR},
      url={https://arxiv.org/abs/2602.16800}, 
}

@misc{digital_ai_at_work_2025,
  title        = {445,000 Public Officers in Malaysia to Benefit from Generative {AI} Under the {AI} at Work 2.0 Initiative by the Ministry of Digital and Google Cloud},
  author       = {{Ministry of Digital Malaysia and Google Cloud}},
  year         = {2025},
  month        = feb,
  howpublished = {\url{https://ai.gov.my/media/news-details/20240205_Press_Release_445000_Public_Officers_in_Malaysia_to_Benefit_from_Generative_AI_Under_the_AI_at_Work_2_Initiative_by_the_Ministry_of_Digital_and_Google_Cloud.pdf}},
  note         = {Accessed 2026-05-01}
}

@misc{jdn_public_sector_ai_guidelines,
  title        = {Guidelines for AI Adoption in the Public Sector},
  author       = {{Jabatan Digital Negara}},
  year         = {2026},
  howpublished = {\url{https://www.jdn.gov.my/en/guidelines-for-the-adoption-of-ai-in-the-public-sector/}},
  note         = {Accessed 2026-05-01}
}

@misc{malaysia_aige,
  title        = {National Guidelines on Artificial Intelligence Governance and Ethics (AIGE)},
  author       = {{Government of Malaysia}},
  year         = {2024},
  howpublished = {\url{https://www.malaysia.gov.my/en/my-initiative/whole-government-digital-services/tadbir-urus/national-guidelines-on-artificial-intelligence-governance-and-ethics-aige}},
  note         = {Accessed 2026-05-01}
}

@misc{edpb_llm_privacy_2025,
  title        = {AI Privacy Risks and Mitigations: Large Language Models},
  author       = {{European Data Protection Board}},
  year         = {2025},
  howpublished = {\url{https://www.edpb.europa.eu/system/files/2025-04/ai-privacy-risks-and-mitigations-in-llms.pdf}},
  note         = {Accessed 2026-05-01}
}

@misc{confidential_prompting,
  title         = {Confidential Prompting: Protecting User Prompts from Cloud {LLM} Providers},
  author        = {Gim, In and Li, Caihua and Zhong, Lin},
  year          = {2024},
  eprint        = {2409.19134},
  archivePrefix = {arXiv},
  primaryClass  = {cs.CR},
  url           = {https://arxiv.org/abs/2409.19134}
}

@misc{ocbc_ibg_account_structure,
  author       = {{OCBC Bank}},
  title        = {Account No. Length Information: IBG Participating Banks Account Structure},
  year         = {n.d.},
  howpublished = {OCBC Bank Malaysia},
  url          = {https://www.ocbc.com.my/assets/pdf/Giro/IBG_Participating_Banks_Account_Structure.pdf},
  note         = {Accessed 2026-09-09}
}
\bibliographystyle{plainnat}

\newpage
\appendix
\onecolumn

\section{Benchmark Construction Details}
\label{app:benchmark-construction}
\subsection{Mapping Malaysian IPUMS Variables}
\label{app:ipums-mapping}

\cref{tab:ipums-mapping} summarizes how we map the indirect identifiers used in RAT-Bench to the Malaysia IPUMS 2000 variables. RAT-Bench is based on U.S. PUMS-style demographic attributes, so several fields do not have exact equivalents in the Malaysian IPUMS extract. We therefore clean, merge, or approximate IPUMS variables to preserve the same broad indirect-identifier categories while adapting them to the Malaysian census schema.

\begin{table}[h]
\caption{Mapping from RAT-Bench/PUMS indirect identifiers to Malaysian IPUMS variables. Date of birth and postcode are added after synthetic microdata generation for transcript generation and re-identification scoring.}
\label{tab:ipums-mapping}
\centering
\small
\begin{tabular}{p{0.13\linewidth}p{0.24\linewidth}p{0.21\linewidth}p{0.30\linewidth}}
\toprule
RAT-Bench field & Meaning in RAT-Bench/PUMS & Malaysia IPUMS field(s) & Adaptation \\
\midrule
ST & U.S. state of residence & BPLS, BPLC & Malaysia IPUMS does not provide current state of residence in our extract. We use Malaysian state of birth as the closest available geographic proxy. For foreign-born records, BPLC provides country of birth instead of a Malaysian state. \\
SEX & Sex & SEX & Used directly. \\
AGE & Age & AGE & Used directly. Following RAT-Bench, we generate date of birth by sampling a valid month and day consistent with calendar constraints, then assigning a birth year from age. Date of birth is generated from age using 1 January 2000 as the reference date. For age values coded as 99+, we randomly assign an age between 99 and 110 before generating date of birth. \\
RAC2P & Race / ethnicity & ETHNIC & Mapped to Malaysian ethnicity categories, including Malay, Chinese, Indian, other Bumiputera, Others, and non-citizens. \\
MAR & Marital status & MARST & Used directly. \\
SCHL & Educational attainment & EDLEV, DEGREE & Merged into a single education variable to better approximate the granularity of PUMS \texttt{SCHL}. We use \texttt{DEGREE} when a valid certificate value exists; otherwise we fall back to \texttt{EDLEV}. \\
ESR & Employment status & WORKING & Used directly. \\
OCCP & Occupation / job title & OCC3 & Used directly. \\
CIT & Citizenship status & CITIZEN1, CITIZEN2 & CITIZEN1 is used as the main citizenship variable. CITIZEN2 contains the country of origin and is a supporting field for non-citizens, used when generating passport numbers. \\
\bottomrule
\end{tabular}
\end{table}

For geography, RAT-Bench uses U.S. state of residence, whereas the Malaysia IPUMS extract available to us provides place of birth. We therefore use Malaysian state of birth as the closest available geographic proxy. Foreign-born individuals do not have a Malaysian birth state, so for these records we use the IPUMS \texttt{CITIZEN2} attribute to obtain country-of-origin information and assign passport-style identifiers rather than NRIC numbers.

Education also requires merging. U.S. PUMS provides a relatively fine-grained educational-attainment variable, \texttt{SCHL}, which distinguishes levels such as kindergarten, individual school grades, high-school completion, some college, bachelor's degree, master's degree, and doctorate. In contrast, Malaysia IPUMS separates education into two less directly comparable fields: \texttt{EDLEV}, which gives a broad highest education level such as no schooling, primary, secondary, post-secondary, or tertiary education; and \texttt{DEGREE}, which gives the highest certificate obtained, such as lower-secondary, upper-secondary, post-secondary, diploma, degree, or postgraduate qualification.

We construct a single education variable by using \texttt{DEGREE} when a valid certificate value is available, and falling back to \texttt{EDLEV} otherwise. This preserves certificate-level detail for individuals with formal qualifications while still retaining broad education information for individuals without a recorded certificate.

\subsection{Synthetic Microdata Generation}
\label{app:synthetic-microdata}

We use Malaysia IPUMS year-2000 microdata, the latest Malaysian IPUMS microdata available to us. Because the original IPUMS records cannot be publicly released under the data-use license, it is only used to construct the privacy-preserving synthetic microdata for this work. After applying the variable cleaning and merging described in \cref{app:ipums-mapping}, we generate synthetic microdata using PrivBayes with privacy budget $\epsilon=5$ via the ReProSyn library \citep{zhang2017privbayes,reprosyn}. We use $\epsilon=5$ as a practical privacy--utility trade-off: the synthetic data must preserve enough demographic dependence structure for re-identification-risk estimation, while avoiding direct release of the underlying restricted microdata.

The resulting synthetic microdata contains 400{,}000 records. Following RAT-Bench, this synthetic population has two roles. First, it serves as the reference population for estimating re-identification risk. Second, it is the pool from which we select benchmark records. For each candidate benchmark case, we sample five indirect identifiers from the available identifier set and estimate the record's baseline correctness under those identifiers. We then select 100 high-risk records whose baseline correctness exceeds $\theta_0=0.9$. This follows the RAT-Bench design choice of evaluating anonymizers on records that are meaningfully re-identifiable before anonymization, rather than on records that would be low-risk even without protection.

\subsection{Date-of-Birth Enrichment}
\label{app:dob}

As with PUMS, date of birth is not directly available in the Malaysia IPUMS extract. We therefore generate date of birth from age.

We assume a reference date of 1 January 2000, matching the census year. For each record, we sample a random birth month and day, taking into account month lengths and leap years, then assign a birth year consistent with the recorded age at the reference date. For rows with that age value of 99+, we sample an age between 99 and 110. The generated date of birth is used both during transcript generation and during re-identification scoring. If the attacker does not recover the full date of birth, correct components such as day, month, or year can still contribute to indirect re-identification risk.

\subsection{Postcode Enrichment}
\label{app:postcode}

The postcode-enrichment procedure is summarized in the main paper. Here, we provide additional implementation details. Malaysia IPUMS does not provide postcode-level geography, so we assign each synthetic record a state-consistent postcode after synthetic microdata generation. The assigned postcode is used both to generate realistic Malaysian addresses and to support re-identification scoring when an attacker recovers a postcode or partial address.

We use district population data from the Malaysia Department of Statistics and postcode data from a public Malaysian postcode repository \citep{opendosm_population_district,malaysia_postcodes}. The population data is from 2020, which is the closest available year to the 2000 IPUMS microdata in the source we use. Although this does not reconstruct year-2000 postcode frequencies exactly, it provides a more grounded assignment than uniformly sampling from all postcodes in a state because postcode selection approximately reflects district-level population distribution.

For each record, we first sample a district \(d\) within state \(s\) proportional to district population:
\[
P(d \mid s)
=
\frac{\mathrm{population}(d)}
{\sum_{\tilde{d} \in \mathcal{D}(s)} \mathrm{population}(\tilde{d})},
\]
where \(\mathcal{D}(s)\) is the set of districts in state \(s\). We then sample a postcode uniformly from the postcode pool matched to that district:
\[
P(z \mid d)
=
\frac{1}{|\mathcal{Z}_d|},
\]
where \(z\) is the sampled postcode and \(\mathcal{Z}_d\) is the set of postcodes matched to district \(d\).

Equivalently, for a postcode \(z\) matched to district \(d_z\), the state-conditional sampling probability is
\[
P(z \mid s)
=
\frac{\mathrm{population}(d_z)}
{\sum_{\tilde{d} \in \mathcal{D}(s)} \mathrm{population}(\tilde{d})}
\cdot
\frac{1}{|\mathcal{Z}_{d_z}|}.
\]

The two data sources use different geographic granularities: the population data is reported at the district level, while the postcode database is organized by city or area. We therefore map postcode entries to districts using name matching followed by manual review of mismatches. If a district cannot be matched to any postcode pool, we exclude it from postcode sampling and renormalize the district-sampling probabilities over the matched districts in the same state. After sampling a district, we sample uniformly from the unique postcode pool matched to that district.

Unlike RAT-Bench's U.S. setting, where ZIP code can be mapped to PUMA geography, our Malaysian setting has no PUMA-equivalent field. We therefore use postcode directly in the re-identification model.

\section{Direct Identifier Generation}
\label{app:direct-identifiers}

After selecting the 100 high-risk benchmark records, we generate Malaysian direct identifiers for each record before transcript generation. RAT-Bench includes U.S.-centric direct identifiers such as Social Security numbers and U.S. addresses. Our Malaysian extension instead uses identifier formats and contextual constraints relevant to Malaysia. We generate nine direct identifier categories: name, email, address, credit card number, phone number, NRIC/passport number, bank account number, EPF number, and TIN.

\paragraph{Coverage balancing.}
Rather than sampling direct identifier types fully at random, we stratify assignment to ensure that all identifier categories are evaluated. In each 100-record explicit setting, eight categories appear 11 times and one category appears 12 times; in our experiments, email appears 12 times. This avoids a split in which locally important identifiers such as NRIC, EPF, TIN, or bank account numbers appear too rarely to evaluate. It also makes the direct-identifier distribution more stable across Easy and Hard settings.

\begin{table}[h]
\caption{Direct identifier categories generated for the Malaysian benchmark. All identifiers are synthetic and used only for benchmark construction.}
\label{tab:direct-identifiers}
\centering
\small
\begin{tabular}{p{0.22\linewidth}p{0.68\linewidth}}
\toprule
Identifier type & Generation approach \\
\midrule
Name & Generated using Claude Sonnet 4.5 with state, ethnicity, gender, citizenship status, and, for non-citizens, country-of-origin context. Prompts cover Malay, Chinese Malaysian, Malaysian Indian, other Bumiputera (non-Malay indigenous groups), mixed-heritage, and foreign-name patterns. \\
Email & Generated from synthetic names using common email-provider domains such as Gmail, Outlook, Hotmail, and Yahoo. \\
Address & Generated with an LLM from Malaysian address-style examples, constrained to the assigned state and postcode. Real address examples are used only as style seeds; final addresses are synthetic. \\
Credit card & Generated using card schemes relevant to Malaysia: Visa, Mastercard, American Express, and UnionPay. Generated numbers pass the Luhn check. \\
Phone number & Generated as mobile or landline numbers. Mobile numbers use Malaysian 01x prefixes; landlines use state-conditioned area codes. \\
NRIC/passport & Malaysian citizens born in Malaysia receive synthetic NRIC numbers consistent with date of birth, state code, and gender. Non-citizens and permanent residents receive passport-style identifiers conditioned on country of origin where possible. \\
Bank account & Generated using account-number lengths from banks operating in Malaysia. \\
EPF & Generated as a unique 8-digit synthetic Employee Provident Fund number. \\
TIN & Generated as an individual Malaysian TIN using the IG prefix followed by 9--12 digits. \\
\bottomrule
\end{tabular}
\end{table}

\subsection{Name Generation}

Names are generated with Claude Sonnet 4.5 using prompts conditioned on the synthetic record's state, ethnicity, gender, citizenship status, and, for non-citizens, country of origin. The goal is to create plausible Malaysian names with realistic variation. We generate names separately for major Malaysian naming contexts: Malay, Chinese Malaysian, Malaysian Indian, other Bumiputera, mixed-heritage, and foreign-origin profiles.

The prompts encode culturally specific naming conventions. Malay names may include patronymics such as ``bin'' and ``binti'', which mean ``son of'' and ``daughter of'', and occasional honorific or hereditary titles. Chinese Malaysian names include both English-first and romanized Chinese forms, with variation in surname and given-name structure. Malaysian Indian names cover several locally common patterns, including names with ``A/L'' and ``A/P'' patronymic markers, where ``A/L'' abbreviates \emph{anak lelaki} (son) and ``A/P'' abbreviates \emph{anak perempuan} (daughter); English-given-name variants combined with Indian name elements; and Sikh/Punjabi-style names using Singh or Kaur where appropriate. Other Bumiputera names are generated with regional guidance for Sabah, Sarawak, and Peninsular Orang Asli contexts. Mixed-heritage names are generated using multiple plausible Malaysian blend patterns, such as Malay--Chinese, Chinese--Indian, Malay--Indian, and Eurasian-influenced forms, rather than repeating a single template. Generated names are checked for uniqueness within the generated pool.

\subsection{Email Addresses}

Email addresses are generated following RAT-Bench's procedure, using the synthetic person's name and generated date of birth. We construct the local-part from common email-address patterns, including first-name and last-name combinations, initials, separators such as periods or underscores, and optional numeric components derived from the birth day, birth month, or birth year. We then assign common email-provider domains such as Gmail, Outlook, Hotmail, and Yahoo. This produces realistic-looking emails whose components are partially correlated with the record, while ensuring that all email addresses are synthetic and unique within the generated pool.

\subsection{Address Generation}

Addresses are generated to reflect Malaysian residential address structure. We first enrich each synthetic record with a state-consistent postcode as described in \cref{app:postcode}. The LLM address generator then produces a synthetic address constrained to the assigned state and postcode.

To capture realistic formatting complexity, we seed the LLM generator, Claude Sonnet 4.5, with examples of Malaysian addresses and instruct it to preserve similar ordering, punctuation, and length without copying any example exactly. Generated addresses may include unit or house numbers, road names, taman or kampung names, districts or towns, postcodes, states, and the country name. The postcode at the end of the generated address is aligned with the postcode stored in the enriched microdata, so that partial address recovery can be scored through postcode correctness.

\subsection{Phone Numbers}

Phone numbers are generated as either mobile or landline numbers. When state information is available, each record receives a mobile number with probability 0.5 and a landline number with probability 0.5. If state information is unavailable, the record receives a mobile number.

Mobile numbers are sampled from Malaysian mobile prefixes, including 010, 011, 012, 013, 014, 016, 017, 018, and 019. The 011 prefix receives a longer subscriber number than the other prefixes. Landline numbers are conditioned on state using Malaysian area-code structure. For example, Kuala Lumpur and Selangor use 03, northern states use 04 or 05 depending on state, and Sabah/Sarawak use 08x prefixes. Generated numbers are validated against formatting rules and de-duplicated.

\subsection{NRIC and Passport Numbers}

For Malaysian citizens born in Malaysia, we generate synthetic NRIC numbers. The first six digits encode the generated date of birth, the middle state code is sampled from valid state-code options for the record's state of birth, and the final digit is sampled to match gender parity conventions. We restrict NRIC assignment to Malaysian citizens born in Malaysia because assigning NRIC state codes to Malaysian citizens born abroad would require additional assumptions. Non-citizens and permanent residents receive passport-style identifiers instead. Passport formats are generated based on country of origin where possible, using country-specific letter and digit patterns.

\subsection{EPF, TIN, and Bank Account Numbers}

We include EPF, TIN, and bank account numbers because these are locally meaningful administrative or financial identifiers. EPF numbers are generated as unique 8-digit synthetic identifiers. Individual TINs are generated using the Malaysian individual ``IG'' prefix followed by 9--12 digits. Bank account numbers are generated by first sampling a bank operating in Malaysia and then sampling an account-number length valid for that bank, using published Malaysian interbank account-structure guidance \citep{ocbc_ibg_account_structure}. The generated bank-account value includes the bank name and a synthetic account number.

\subsection{Credit Card Numbers}

Credit card numbers are included for comparability with RAT-Bench, but we adapt the card-scheme set to the Malaysian context. RAT-Bench includes schemes such as Visa, Mastercard, American Express, Discover, Diners Club, and JCB. We instead generate Visa, Mastercard, American Express, and UnionPay numbers. Synthetic card numbers are generated with scheme-appropriate prefixes and lengths and are required to pass the Luhn check. No real card numbers are used.

\subsection{Stratified direct identifier assignment.}
To ensure that all direct identifier types are represented and comparable across difficulty settings, we stratify direct identifier assignment. In each 100-record explicit setting, each of the nine direct identifier categories appears 11 times, with the email category appearing 12 times.

\section{Text Generation and Quality Control}
\label{app:text-generation}

We follow RAT-Bench in generating transcripts for three scenarios: medical consultation, chatbot conversation, and meeting transcript. We create two localized language settings using separate prompts: Malaysian English and Bahasa Malaysia.

Malaysian English refers to English dialogue grounded in Malaysian context, including Malaysian names, geography, institutions, social references, and light local phrasing where natural. Bahasa Malaysia refers to Malay-language dialogue, with light English code-switching where natural.

We use the same three difficulty settings as RAT-Bench: Easy, Hard, and Implicit. In the Easy setting, target identifiers are stated explicitly in standard form. In the Hard setting, identifiers are still explicit, but appear in less standard forms such as spelled-out numbers, segmentation, slang, abbreviations, informal formatting, or mixed-language phrasing. In the Implicit setting, target values are not stated directly; instead, indirect identifiers must be inferred from Malaysian contextual clues such as geography, cultural practices, routines, institutions, and social references.

\subsection{Implicit Generation Quality Control}
\label{app:implicit-qc}

Implicit generation required additional quality control. The initial Malaysian English Implicit generations sounded natural but did not always make the target attributes recoverable, giving a no-anonymization re-identification rate of only 5\%. We improved the Implicit-generation prompt and manually corrected 61/100 Malaysian English Implicit examples. After this correction, the no-anonymization re-identification rate increased to 35\%, close to the original RAT-Bench Implicit setting. The resulting Malaysian English Implicit results are reported in \cref{tab:main-results}.

We did not apply the same manual correction step to Bahasa Malaysia Implicit examples. The no-anonymization result for Bahasa Malaysia Implicit remains low at 7\%, so we treat Bahasa Malaysia Implicit as exploratory and omit it from the main comparison. The full Bahasa Malaysia Implicit results are reported separately in \cref{tab:bm-implicit}.

\section{Experimental Details}
\label{app:experimental-details}

\subsection{Anonymizers}
\label{app:anonymizers}

We evaluate most anonymizers from RAT-Bench and add OpenAI Privacy Filter. The evaluated methods are Presidio, Scrubadub, GliNER, UniNER, TEM with $\epsilon=11$, OpenAI Privacy Filter, and two LLM anonymizers using the original RAT-Bench PII Purifier and Rescriber prompts. For the LLM anonymizers, we use Claude Haiku 4.5. We do not modify the RAT-Bench LLM anonymization prompts for the Malaysian setting; this lets us test whether the prompt generalizes to local identifiers and language settings out of the box.

\subsection{Generator and Attacker}
\label{app:generator-attacker}

We use Claude Sonnet 4.5 as the transcript generator and attacker. The generator receives a scenario, target profile, difficulty setting, and attribute examples. The attacker receives the anonymized transcript and is asked to infer target attributes in a structured format.

The attacker prompt follows RAT-Bench but adds two pieces of context: inferences should be made from a Malaysian context, and time-based reasoning should assume the reference date 1 January 2000. The reference date is needed because age and date-of-birth clues are generated relative to the year-2000 microdata.

\subsection{Re-identification Metric}
\label{app:reid-metric}

Following RAT-Bench, direct and indirect identifiers are scored differently. If the attacker correctly recovers at least one direct identifier, the record is treated as re-identified. If no direct identifier is recovered, re-identification risk is computed from the correctly recovered indirect identifiers against the synthetic reference microdata.

We use the same final success threshold as RAT-Bench: a record is counted as re-identified when the estimated re-identification risk exceeds $\theta=0.2$, corresponding to $k=5$. Utility is measured using BLEU between the original transcript and anonymized transcript.

\section{Additional Results}
\label{app:additional-results}

\subsection{Qualitative Bahasa Malaysia NER Failure Example}
\label{app:bm-failure-example}

\begin{figure}[t]
    \centering
    \includegraphics[width=0.95\linewidth]{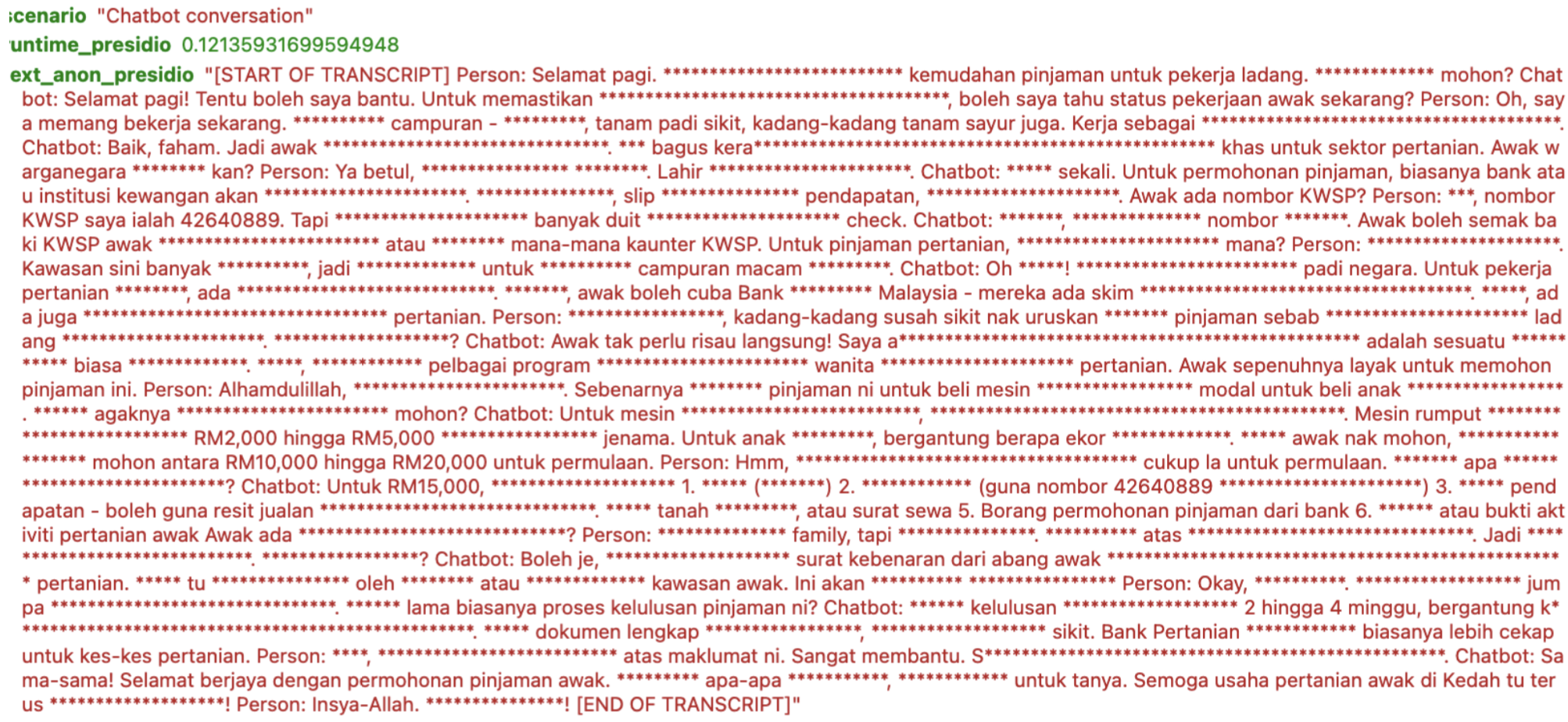}

    \caption{Example of a Bahasa Malaysia anonymization failure. Presidio removes large portions of benign text but leaves the EPF/KWSP number (synthetic) , a local direct identifier, unredacted.}

    \label{fig:bm-anonymization-failure}
\end{figure}

\cref{fig:bm-anonymization-failure} shows an example of the mixed failure mode observed in Bahasa Malaysia. Presidio removes large portions of benign text, reducing utility, but still leaves a local direct identifier exposed. This illustrates why lower re-identification risk should be interpreted together with utility: some NER-based methods reduce risk partly through broad over-redaction rather than precise removal of identifying information.

\subsection{Attribute-Level Privacy Heatmaps}
\label{app:attribute-heatmaps}

To better understand which attributes are protected or exposed, we report two families of heatmaps for both Malaysian English and BM.

\begin{figure}[p]
    \centering

    \begin{subfigure}{0.98\textwidth}
        \centering
        \includegraphics[width=\linewidth,trim={0 0 0 0.5in},clip,height=0.25\textheight,keepaspectratio]{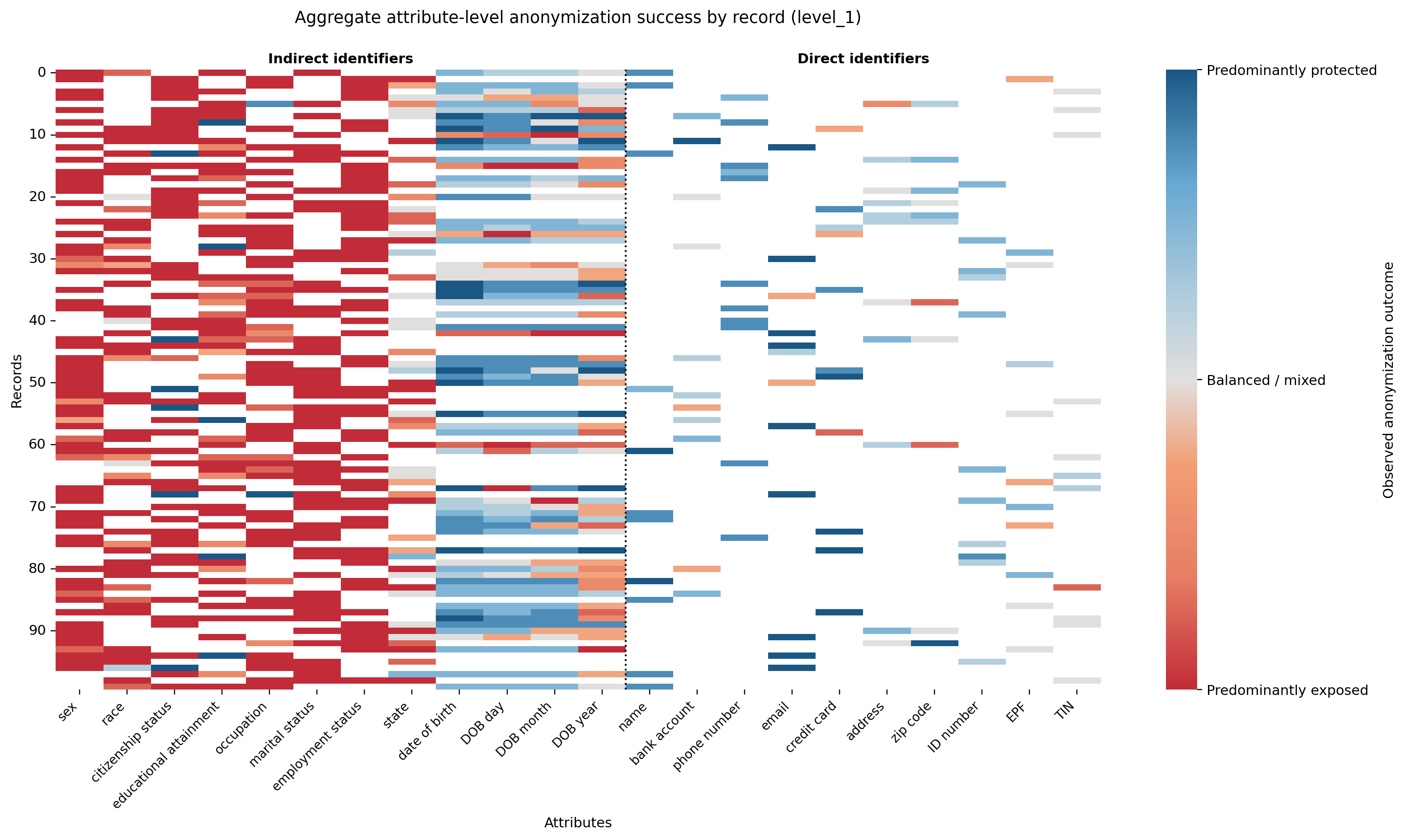}
        \caption{Malaysian English Easy}
        \label{fig:english-easy-record-attribute-privacy}
    \end{subfigure}

    \vspace{0.5em}

    \begin{subfigure}{0.98\textwidth}
        \centering
        \includegraphics[width=\linewidth,trim={0 0 0 0.5in},clip,height=0.25\textheight,keepaspectratio]{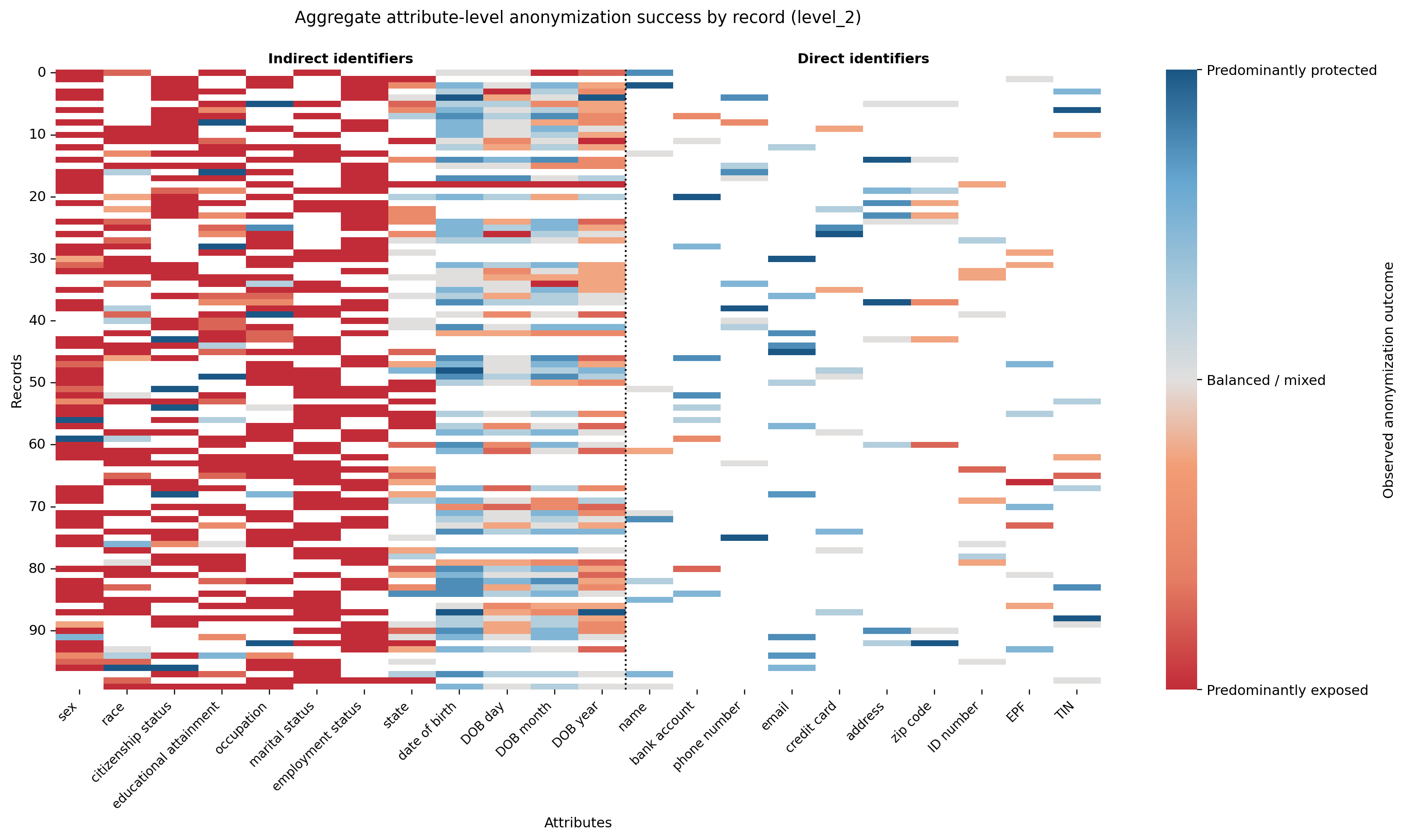}
        \caption{Malaysian English Hard}
        \label{fig:english-hard-record-attribute-privacy}
    \end{subfigure}

    \vspace{0.5em}

    \begin{subfigure}{0.98\textwidth}
        \centering
        \includegraphics[width=\linewidth,trim={0 0 0 0.5in},clip,height=0.25\textheight,keepaspectratio]{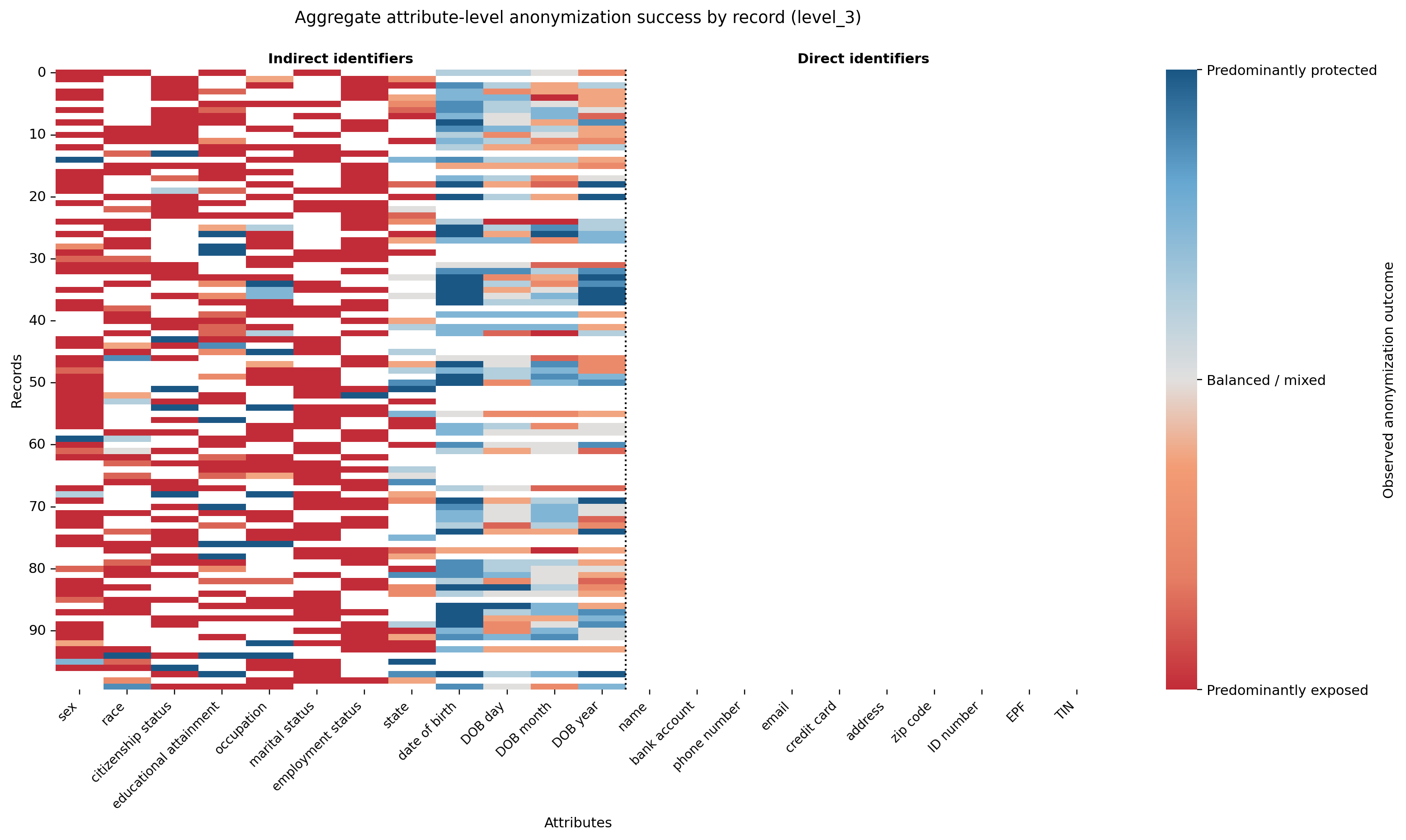}
        \caption{Malaysian English Implicit}
        \label{fig:english-implicit-record-attribute-privacy}
    \end{subfigure}

    \caption{Malaysian English record-by-attribute privacy heatmaps across Easy, Hard, and Implicit settings. Each row is a benchmark record and each column is an identifier. Cell values aggregate privacy success across anonymizers for that record--identifier pair, where privacy success is defined as \(1-\text{attacker correctness}\). Blue indicates that the identifier was usually protected, red indicates that it was often recovered, and white indicates that the identifier was not present or not targeted for that record. The dotted vertical line separates indirect identifiers from direct identifiers.}
    \label{fig:english-record-attribute-privacy}
\end{figure}

\begin{figure}[p]
    \centering

    \begin{subfigure}{0.98\textwidth}
        \centering
        \includegraphics[width=\linewidth,trim={0 0 0 0.5in},clip,height=0.25\textheight,keepaspectratio]{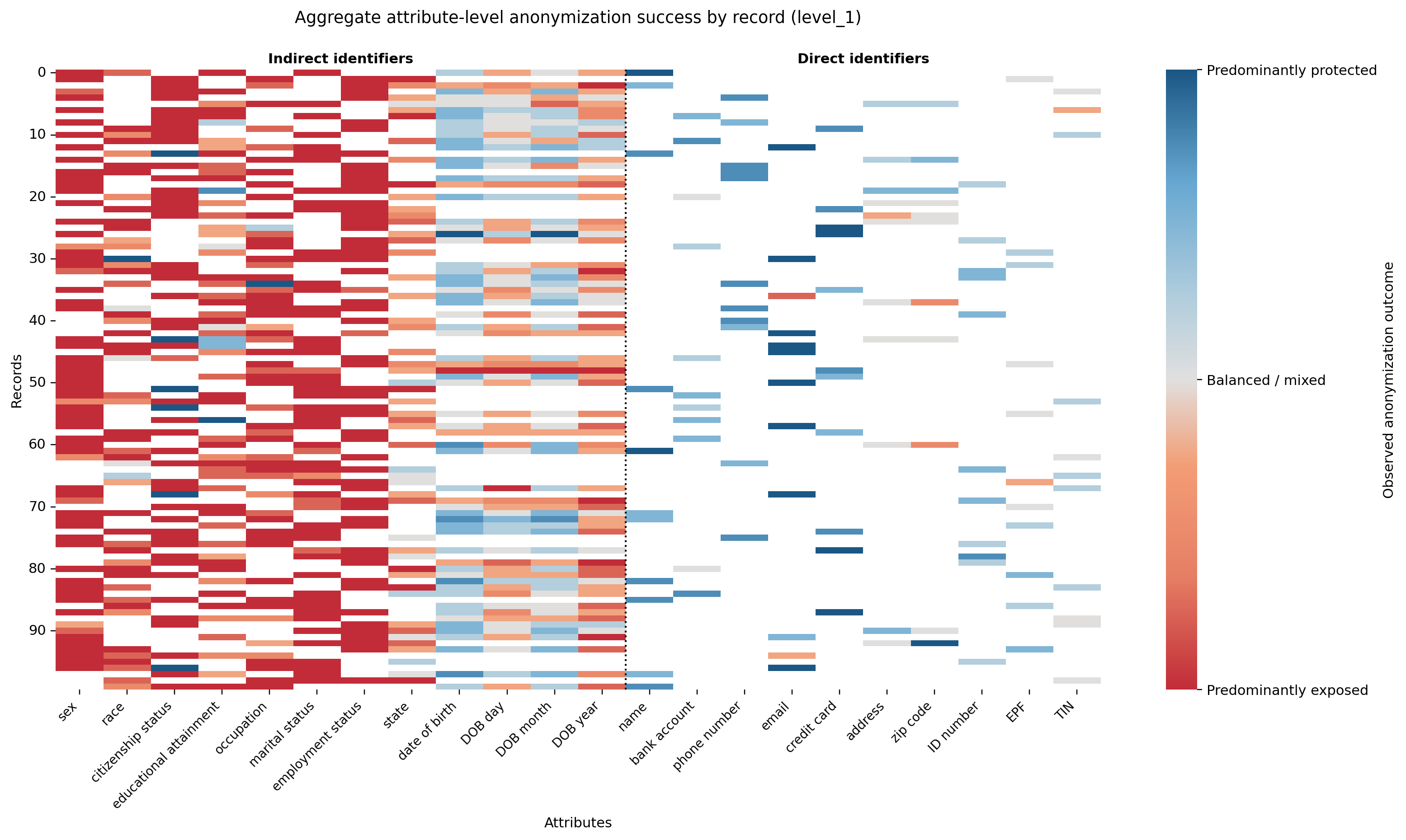}
        \caption{Bahasa Malaysia Easy}
        \label{fig:bm-easy-record-attribute-privacy}
    \end{subfigure}

    \vspace{0.5em}

    \begin{subfigure}{0.98\textwidth}
        \centering
        \includegraphics[width=\linewidth,trim={0 0 0 0.5in},clip,height=0.25\textheight,keepaspectratio]{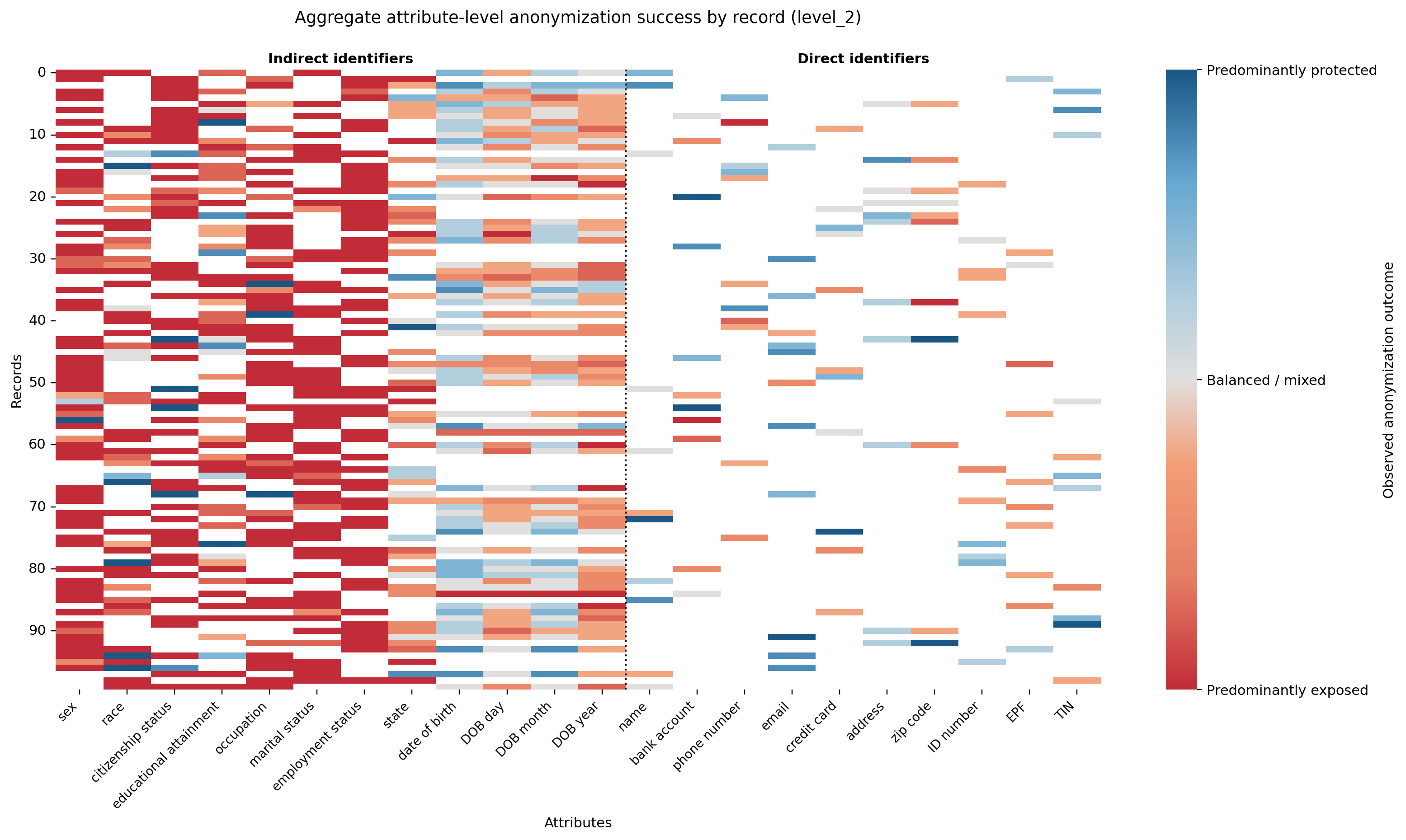}
        \caption{Bahasa Malaysia Hard}
        \label{fig:bm-hard-record-attribute-privacy}
    \end{subfigure}

    \vspace{0.5em}

    \begin{subfigure}{0.98\textwidth}
        \centering
        \includegraphics[width=\linewidth,trim={0 0 0 0.5in},clip,height=0.25\textheight,keepaspectratio]{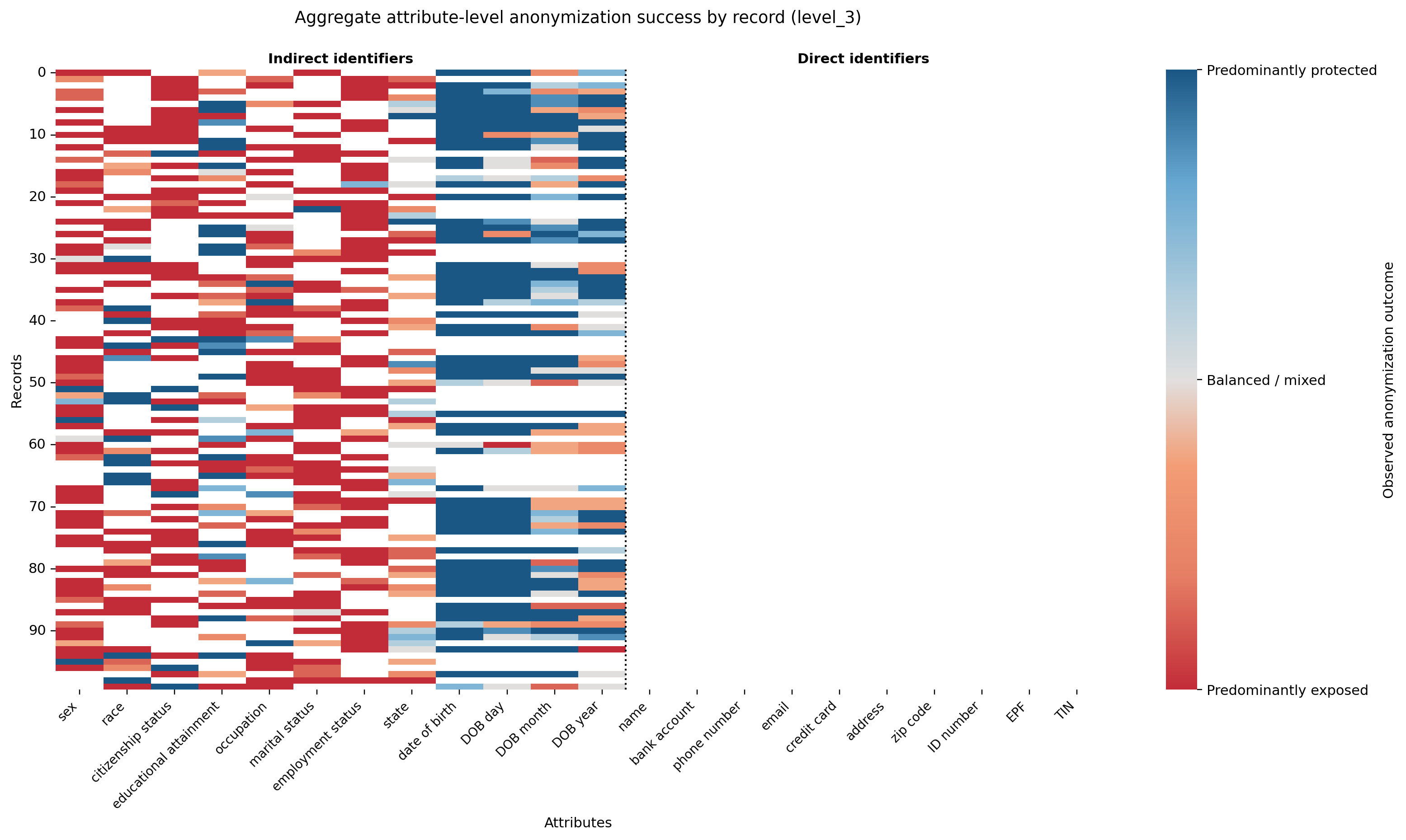}
        \caption{Bahasa Malaysia Implicit}
        \label{fig:bm-implicit-record-attribute-privacy}
    \end{subfigure}

    \caption{Bahasa Malaysia record-by-attribute privacy heatmaps across Easy, Hard, and Implicit settings. Each row is a benchmark record and each column is an identifier. Cell values aggregate privacy success across anonymizers for that record--identifier pair, where privacy success is defined as \(1-\text{attacker correctness}\). Blue indicates that the identifier was usually protected, red indicates that it was often recovered, and white indicates that the identifier was not present or not targeted for that record. The dotted vertical line separates indirect identifiers from direct identifiers.}
    \label{fig:bm-record-attribute-privacy}
\end{figure}

The first family of heatmaps, \cref{fig:english-record-attribute-privacy,fig:bm-record-attribute-privacy}, is record-by-attribute. Each row corresponds to one benchmark record, and each column corresponds to an indirect or direct identifier. Cell values aggregate privacy success across anonymizers for that record--attribute pair, where privacy success is defined as \(1 - \text{attacker correctness}\). Blue cells indicate that the attribute was usually protected from the attacker, while red cells indicate that the attribute was often recovered. White cells indicate that the attribute was not present or not targeted for that record. This is expected because each benchmark record contains only a subset of indirect identifiers and, in the Easy and Hard settings, one direct identifier. For date of birth, we follow RAT-Bench's partial-credit scoring: if the full date of birth is not recovered, day, month, and year are evaluated separately.

\begin{figure}[p]
    \centering

    \begin{subfigure}{0.98\textwidth}
        \centering
        \includegraphics[width=\linewidth,trim={0 0 0 0.5in},clip,height=0.25\textheight,keepaspectratio]{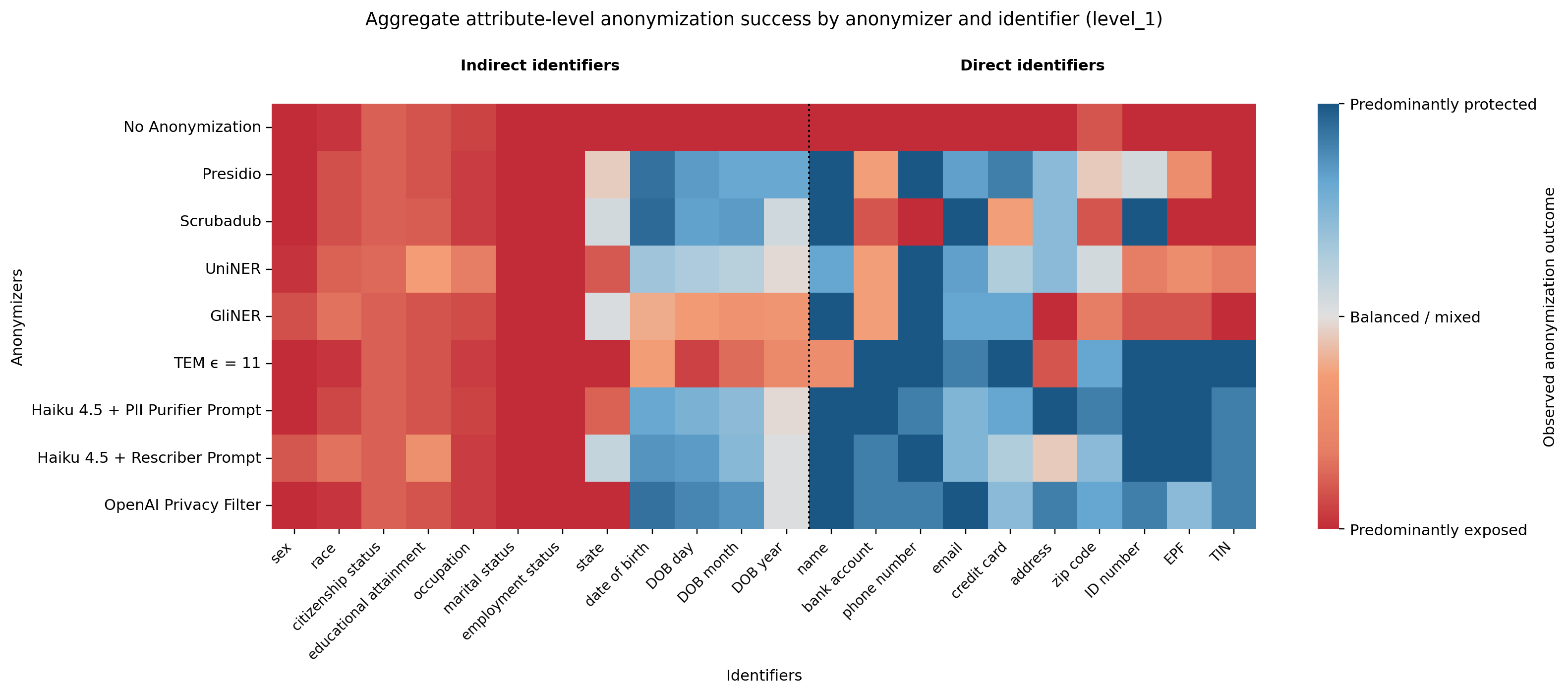}
        \caption{Malaysian English Easy}
        \label{fig:english-easy-anonymizer-identifier-privacy}
    \end{subfigure}

    \vspace{0.5em}

    \begin{subfigure}{0.98\textwidth}
        \centering
        \includegraphics[width=\linewidth,trim={0 0 0 0.5in},clip,height=0.25\textheight,keepaspectratio]{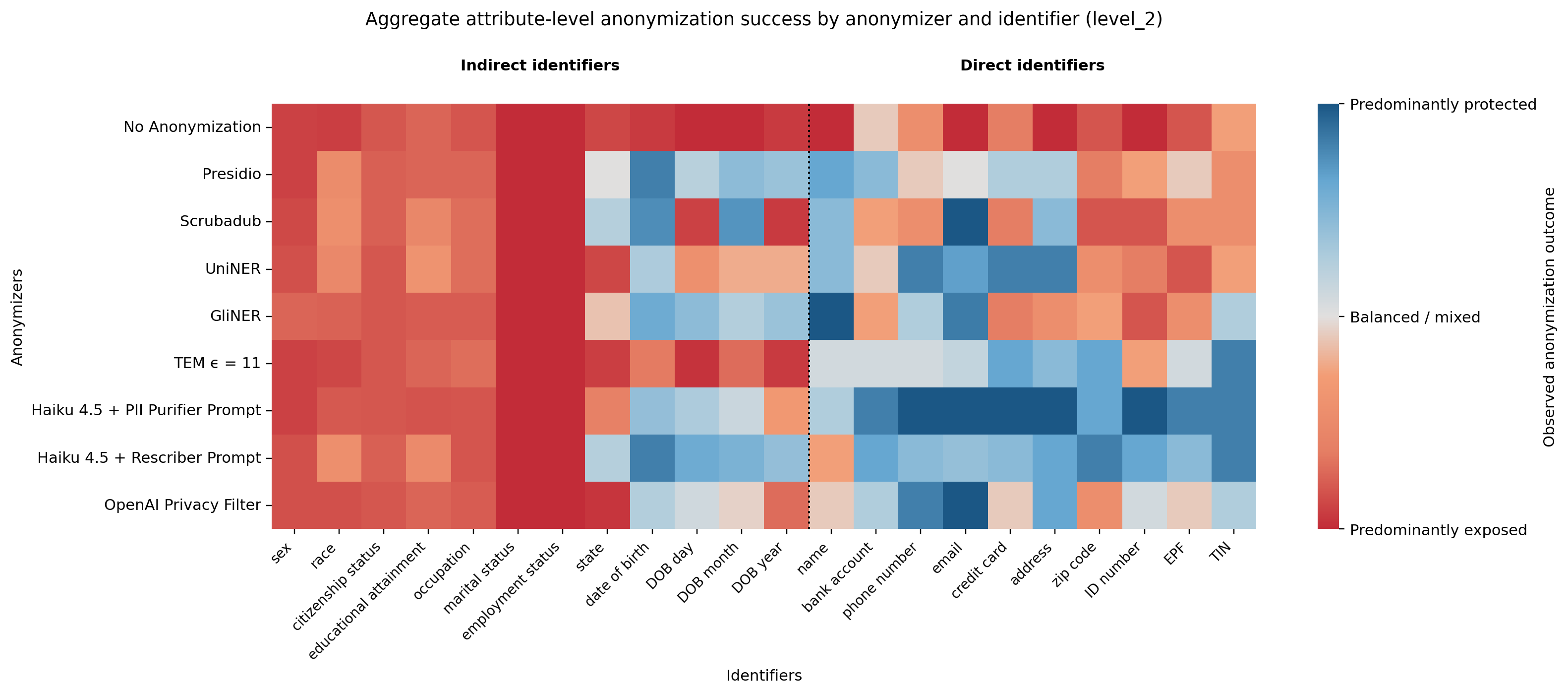}
        \caption{Malaysian English Hard}
        \label{fig:english-hard-anonymizer-identifier-privacy}
    \end{subfigure}

    \vspace{0.5em}

    \begin{subfigure}{0.98\textwidth}
        \centering
        \includegraphics[width=\linewidth,trim={0 0 0 0.5in},clip,height=0.25\textheight,keepaspectratio]{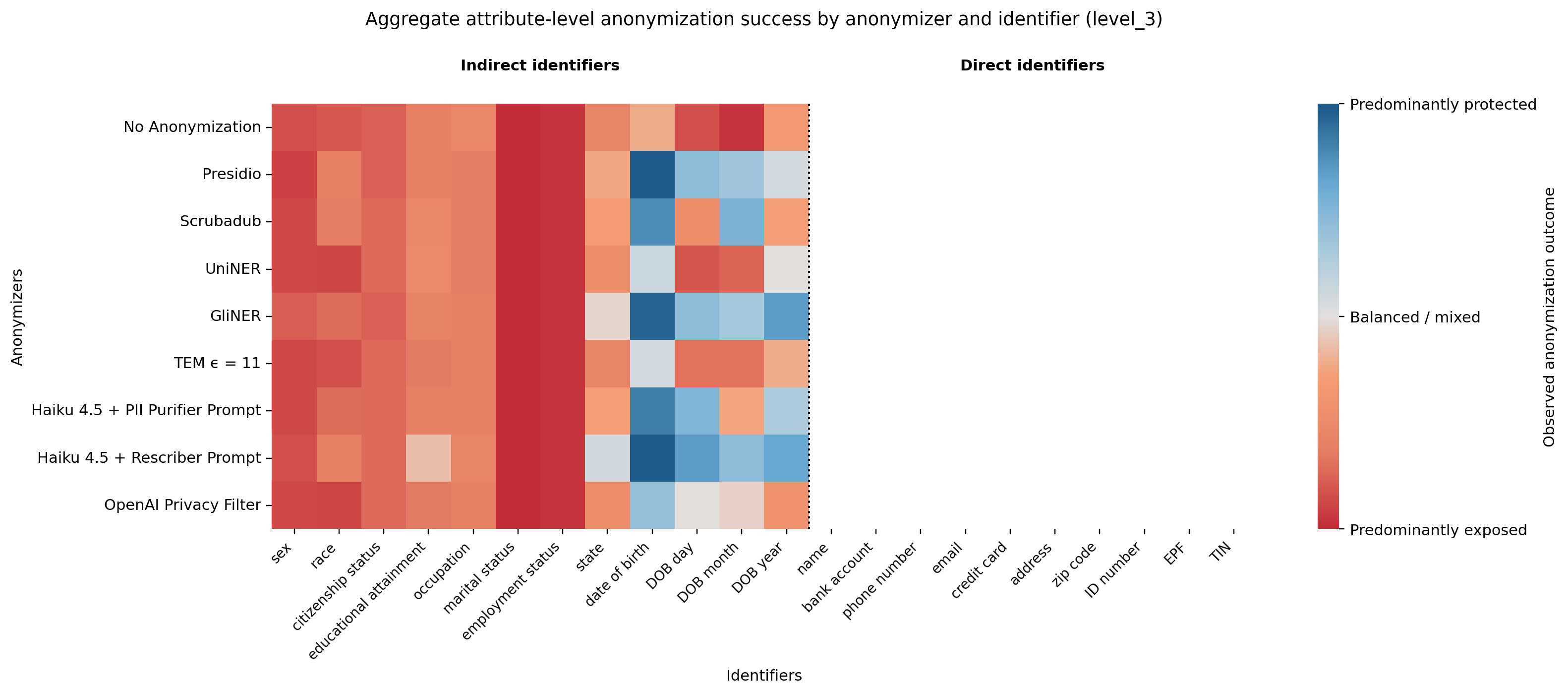}
        \caption{Malaysian English Implicit}
        \label{fig:english-implicit-anonymizer-identifier-privacy}
    \end{subfigure}

    \caption{Malaysian English anonymizer-by-identifier privacy heatmaps across Easy, Hard, and Implicit settings. Each row is an anonymization method and each column is an identifier. Cell values show mean privacy success across records for that anonymizer--identifier pair, where privacy success is defined as \(1-\text{attacker correctness}\). Blue indicates stronger protection and red indicates weaker protection. The dotted vertical line separates indirect identifiers from direct identifiers.}
    \label{fig:english-anonymizer-identifier-privacy}
\end{figure}

\begin{figure}[p]
    \centering

    \begin{subfigure}{0.98\textwidth}
        \centering
        \includegraphics[width=\linewidth,trim={0 0 0 0.5in},clip,height=0.25\textheight,keepaspectratio]{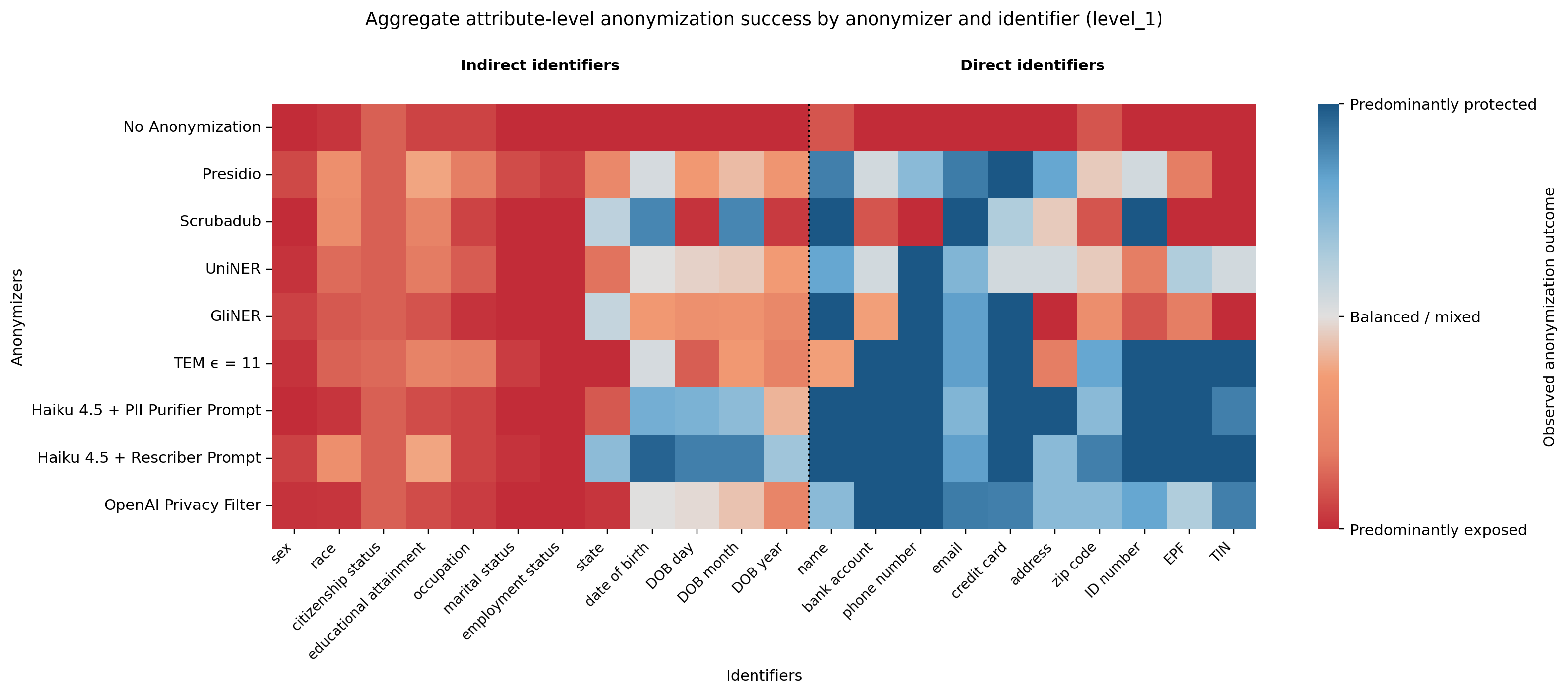}
        \caption{Bahasa Malaysia Easy}
        \label{fig:bm-easy-anonymizer-identifier-privacy}
    \end{subfigure}

    \vspace{0.5em}

    \begin{subfigure}{0.98\textwidth}
        \centering
        \includegraphics[width=\linewidth,trim={0 0 0 0.5in},clip,height=0.25\textheight,keepaspectratio]{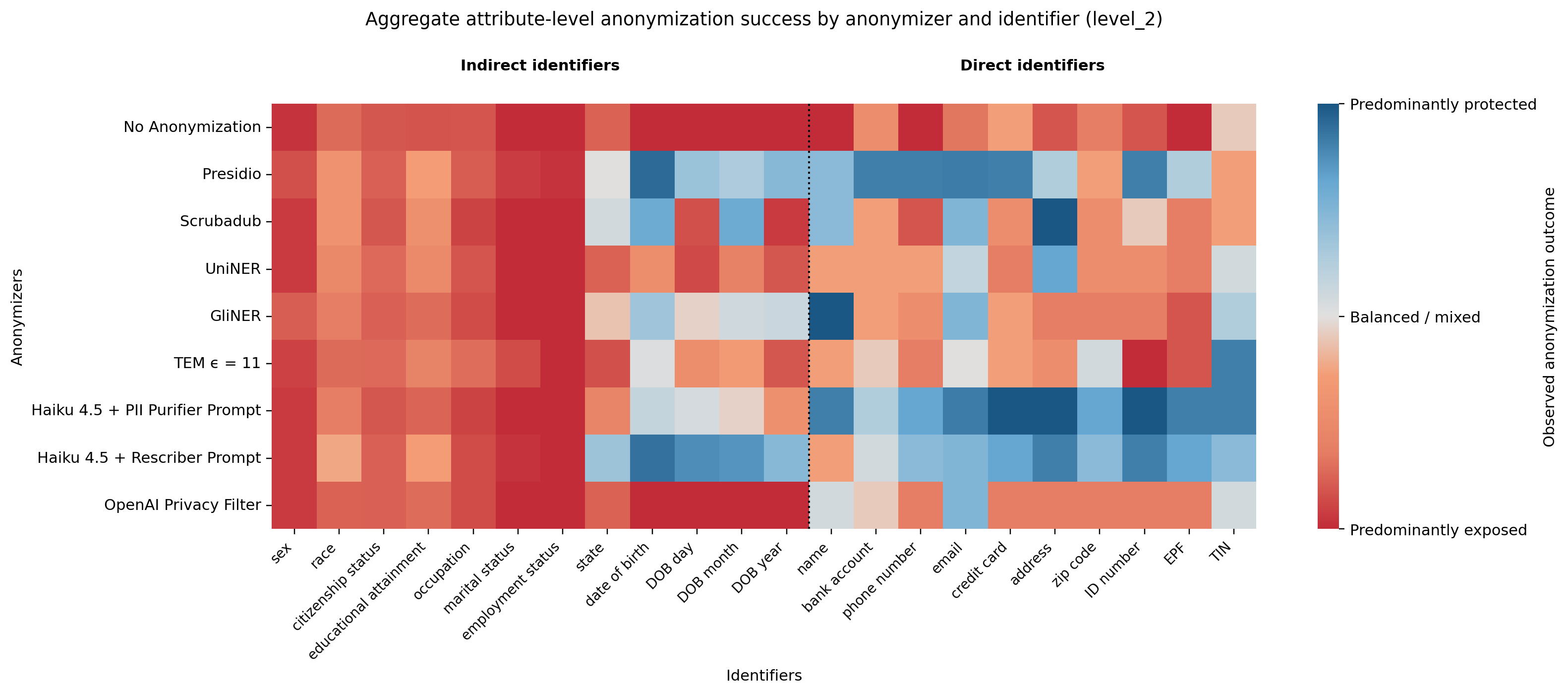}
        \caption{Bahasa Malaysia Hard}
        \label{fig:bm-hard-anonymizer-identifier-privacy}
    \end{subfigure}

    \vspace{0.5em}

    \begin{subfigure}{0.98\textwidth}
        \centering
        \includegraphics[width=\linewidth,trim={0 0 0 0.5in},clip,height=0.25\textheight,keepaspectratio]{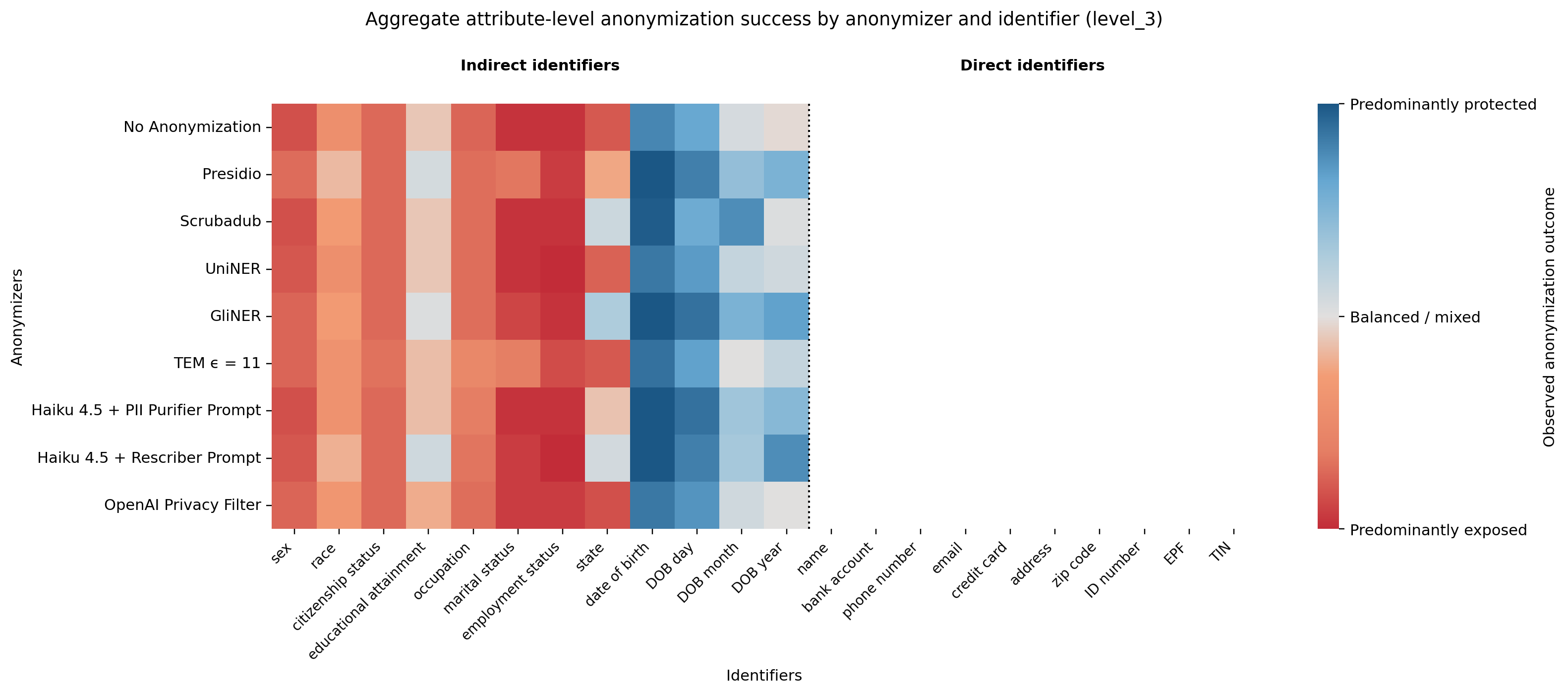}
        \caption{Bahasa Malaysia Implicit}
        \label{fig:bm-implicit-anonymizer-identifier-privacy}
    \end{subfigure}

    \caption{Bahasa Malaysia anonymizer-by-identifier privacy heatmaps across Easy, Hard, and Implicit settings. Each row is an anonymization method and each column is an identifier. Cell values show mean privacy success across records for that anonymizer--identifier pair, where privacy success is defined as \(1-\text{attacker correctness}\). Blue indicates stronger protection and red indicates weaker protection. The dotted vertical line separates indirect identifiers from direct identifiers.}
    \label{fig:bm-anonymizer-identifier-privacy}
\end{figure}

The second family of heatmaps, \cref{fig:english-anonymizer-identifier-privacy,fig:bm-anonymizer-identifier-privacy}, is anonymizer-by-identifier. Each row corresponds to an anonymization method, and each column corresponds to an identifier. Cell values show the mean privacy success for that anonymizer--identifier pair across all records in which the identifier appears. These plots show which anonymizers are stronger or weaker for each identifier type.

Across the record-by-attribute heatmaps, indirect identifiers are generally more exposed than direct identifiers, shown by the concentration of red cells on the indirect-identifier side. This is expected because many indirect identifiers are not simple spans to remove; they can remain inferable from context even after anonymization. Direct identifiers such as names, dates of birth, and emails are more consistently protected, but LLM-based methods are observed to perform better on Malaysia-specific identifiers such as EPF and TIN.

The anonymizer-by-identifier heatmaps show that LLM-based anonymizers and the OpenAI Privacy Filter provide stronger protection for many direct identifiers in the Easy setting, while NER-based systems are less consistent across local identifier types. In the Hard setting, where identifiers appear in non-standard forms, the LLM-based methods remain more consistent, while several non-LLM systems show weaker or more uneven protection. These diagnostic plots support the aggregate results in \cref{tab:main-results}: anonymizer performance is not uniform across identifier types, and localized identifiers expose failure modes that are less visible in aggregate re-identification rates alone.

\subsection{Bahasa Malaysia Implicit Results}
\label{app:bm-implicit}

\cref{tab:bm-implicit} reports Bahasa Malaysia Implicit results for completeness. We do not use these results for the main claims because the no-anonymization result is only 7\%, indicating that the Bahasa Malaysia Implicit examples are not yet consistently recoverable.

\begin{table}[h]
\caption{Bahasa Malaysia Implicit results. These are exploratory because the no-anonymization re-identification result is low.}
\label{tab:bm-implicit}
\centering
\begin{tabular}{lrr}
\toprule
Method & $R_{succ}$ (\% $\downarrow$) & BLEU ($\uparrow$) \\
\midrule
No anonymization & 7 & 1.00 \\
Presidio & 1 & 0.20 \\
Scrubadub & 0 & 0.49 \\
GliNER & 0 & 0.52 \\
UniNER & 5 & 0.89 \\
TEM $\epsilon=11$ & 5 & 0.32 \\
Haiku + PII Purifier & 1 & 0.96 \\
Haiku + Rescriber & 0 & 0.84 \\
Privacy Filter & 7 & 0.99 \\
\bottomrule
\end{tabular}
\end{table}

\subsection{Full BLEU Results}
\label{app:full-bleu}

\cref{tab:full-bleu} reports BLEU scores by language and difficulty. The main paper reports average BLEU over Easy and Hard settings, with Implicit BLEU shown separately where relevant.

\begin{table}[h]
\caption{Full BLEU scores by language and difficulty.}
\label{tab:full-bleu}
\centering
\begin{tabular}{lrrrrrrrr}
\toprule
\multirow{2}{*}{Method} & \multicolumn{4}{c}{Malaysian English} & \multicolumn{4}{c}{BM} \\
\cmidrule(lr){2-5}\cmidrule(lr){6-9}
& Easy & Hard & Avg. & Impl. & Easy & Hard & Avg. & Impl. \\
\midrule
No anonymization & 1.00 & 1.00 & 1.00 & 1.00 & 1.00 & 1.00 & 1.00 & 1.00 \\
Presidio & 0.81 & 0.83 & 0.82 & 0.85 & 0.16 & 0.18 & 0.17 & 0.20 \\
Scrubadub & 0.76 & 0.77 & 0.77 & 0.82 & 0.39 & 0.42 & 0.41 & 0.49 \\
GliNER & 0.86 & 0.88 & 0.87 & 0.66 & 0.48 & 0.49 & 0.49 & 0.52 \\
UniNER & 0.65 & 0.64 & 0.65 & 0.96 & 0.86 & 0.87 & 0.87 & 0.89 \\
TEM $\epsilon=11$ & 0.64 & 0.65 & 0.65 & 0.70 & 0.23 & 0.28 & 0.25 & 0.32 \\
Haiku + PII Purifier & 0.94 & 0.94 & 0.94 & 0.97 & 0.93 & 0.91 & 0.92 & 0.96 \\
Haiku + Rescriber & 0.83 & 0.83 & 0.83 & 0.87 & 0.79 & 0.75 & 0.77 & 0.84 \\
Privacy Filter & 0.94 & 0.94 & 0.94 & 0.98 & 0.94 & 0.94 & 0.94 & 0.99 \\
\bottomrule
\end{tabular}
\end{table}

\subsection{Qualitative Comparison with RAT-Bench}
\label{app:ratbench-comparison}

The original RAT-Bench results show the same broad pattern that LLM-based anonymizers tend to occupy a stronger privacy--utility region than many NER-based tools. Our Malaysian results preserve this trend but reveal sharper local failure modes: non-LLM systems may miss Malaysian-specific identifiers, degrade on hard local forms, or over-redact Bahasa Malaysia text.

\section{Prompt Templates}
\label{app:prompts}

\subsection{Generator Prompt}
\label{app:generator-prompt}

The transcript generator prompt is the largest prompt change from RAT-Bench. We retain the RAT-Bench structure of conditioning generation on a scenario, target profile, target attributes, and difficulty setting, but adapt the prompt to Malaysian demographic, linguistic, and cultural context. The prompt is assembled from five components: (i) a scenario description, (ii) the target profile and target attributes, (iii) attribute-specific few-shot examples for the requested difficulty setting, (iv) difficulty-specific instructions, and (v) Malaysian realism requirements. For Bahasa Malaysia generation, the attribute-specific few-shot examples are translated into Bahasa Malaysia, and we append an additional instruction requiring the full conversation to be generated in Bahasa Malaysia with target attributes adapted naturally to that linguistic context. For Implicit examples, the prompt also encourages natural Malaysian contextual clues, including colloquial Bahasa Malaysia, local institutions, geography, festivals, routines, and culturally grounded references where appropriate.

The generator is used for three scenarios: medical consultation, chatbot conversation, and meeting transcript. The output format is constrained to alternating dialogue lines, with fixed speaker labels for each scenario. We use longer word limits for Implicit examples because the target attributes must be recoverable from contextual clue bundles rather than directly stated values.

\paragraph{Difficulty-specific generation instructions.}
The Easy setting requires target values to be stated explicitly in standard form. The Hard setting requires the same target values to remain explicit, but expressed in less standard forms, such as spelled-out numbers, segmented identifiers, abbreviations, colloquial phrasing, unusual formatting, or mixed-language wording. The Implicit setting forbids directly stating target values. Instead, each target attribute must be recoverable from a natural clue bundle, where a clue bundle is either one strongly identifying contextual clue or multiple weaker clues that jointly identify the intended value. For example, state may be implied through hometown, local geography, transport routes, or region-specific references; ethnicity may be implied through natural cultural or festive references; and age or date of birth may be implied through simple date and timeline references.

\paragraph{Malaysian realism instructions.}
The prompt instructs the generator to write dialogue that sounds realistic in Malaysia, using Malaysian English or Bahasa Malaysia as appropriate. It allows light code-switching and local terms when natural, but discourages caricatured language or excessive decorative local detail. Suggested Malaysian context cues include transport systems such as Grab, LRT, MRT, KTM, and balik kampung travel; food and social settings such as mamak, kopitiam, pasar, kenduri, and open houses; and institutional references such as SPM, MyKad, IC, clinic forms, HR forms, and government counters. These cues are used only when they support the scenario or help make the target attributes recoverable.

\paragraph{Attribute-specific examples.}
The prompt includes attribute-specific examples at the requested difficulty setting. These examples are not copied verbatim; they guide the generator toward the expected surface form or clue style. Examples cover both direct identifiers and indirect demographic attributes. For direct identifiers, Easy examples state values in standard form, while Hard examples use non-standard forms such as spelling, segmentation, or informal phrasing. For indirect attributes, Implicit examples use Malaysian contextual clues. For instance, state may be implied through references to places such as Ipoh, Alor Setar, Seremban, Georgetown, Kota Bharu, or KK (Kota Kinabalu); ethnicity may be implied through natural references to Hari Raya, Chinese New Year, Deepavali, Gawai, Kaamatan, or Eurasian family practices; and marital status may be implied through references to checking plans with a spouse, giving angpao as a married adult, living alone after a spouse's death, or adjusting after divorce.

A simplified version of the generator prompt template is shown below:

\begin{quote}
\small\sloppy
\texttt{<SCENARIO>} \\
\texttt{The TARGET ATTRIBUTES provided for the individual must be recoverable from the text at the specified DIFFICULTY LEVEL. The exchange must feel natural, realistic, and grounded in Malaysia. Use the examples as guidance, but do not copy them verbatim.} \\

\texttt{TARGET ATTRIBUTES: <PROFILE>} \\
\texttt{DIFFICULTY LEVEL: <LEVEL>} \\
\texttt{EXAMPLES: <ATTRIBUTE-SPECIFIC EXAMPLES>} \\

\texttt{DIFFICULTY-SPECIFIC REQUIREMENTS: <EASY/HARD/IMPLICIT INSTRUCTIONS>} \\

\texttt{MALAYSIAN REALISM REQUIREMENTS: Use natural Malaysian English or Bahasa Malaysia where appropriate. Light code-switching and local context are allowed when natural. Prefer ordinary Malaysian life details over obscure or overly decorative references.} \\

\texttt{<SCENARIO FORMATTING>} \\
\texttt{The exchange should contain <WORD LIMIT> words and all target attributes should be recoverable at the requested difficulty level.}
\end{quote}

For Bahasa Malaysia generation, the following language instruction is appended:

\begin{quote}
\small\sloppy
\texttt{The conversation must be generated in BM. Although the target attributes, their values, and examples are provided in English, they should be translated and adapted naturally to BM, including the requested difficulty level.}
\end{quote}

\subsection{Anonymizer Prompt}
\label{app:anonymizer-prompt}

The LLM anonymizers use the original RAT-Bench PII Purifier and Rescriber prompts without Malaysia-specific modifications. This design choice lets us test whether the original anonymization prompts generalize out of the box to Malaysian identifiers, Malaysian English, and BM.

\subsection{Attacker Prompt}
\label{app:attacker-prompt}

The attacker prompt follows RAT-Bench and asks the model to infer target attributes from the anonymized transcript in a structured format. We make only one localization change: we add the instruction, ``Make all inferences from a Malaysian context. Assume today's date is 1 January 2000 for any time-based reasoning.'' The Malaysian-context instruction is needed because many implicit clues depend on local geography, naming conventions, festivals, institutions, and administrative identifiers. The reference-date instruction is needed because age and date-of-birth clues are generated relative to the year-2000 microdata.

\end{document}